\documentclass[11pt]{article}

\usepackage[final]{acl}

\usepackage{times}
\usepackage{latexsym}
\usepackage{amsmath} 
\usepackage[T1]{fontenc}

\usepackage[utf8]{inputenc}

\usepackage{microtype}

\usepackage{inconsolata}
\usepackage{multirow}
\usepackage{graphicx}
\usepackage{booktabs}
\usepackage{enumitem}
\usepackage{easyReview}
\usepackage{graphicx}
\usepackage{longtable}
\usepackage[normalem]{ulem}
\newcommand{\ours}{\textsc{ProChatIP}}
\title{Towards Proactive Information Probing: \\Customer Service Chatbots Harvesting Value from Conversation}

\author{
Chen Huang$^{1}$ \quad Zitan Jiang$^{23}$ \quad Changyi Zou$^{23}$ \quad {Wenqiang Lei}$^{23}$\thanks{Corresponding author.} \quad {See-Kiong Ng}$^{1}$ \\
$^{1}$ Institute of Data Science, National University of Singapore  \\
$^{2}$ College of Computer Science, Sichuan University  \\
$^{3}$ Engineering Research Center of Machine Learning and Industry Intelligence, \\ Ministry of Education, China\\
\texttt{huang\_chen@nus.edu.sg, jzitan@stu.scu.edu.cn}
}

\begin{document}
\maketitle
\begin{abstract}
Customer service chatbots are increasingly expected to serve not merely as reactive support tools for users, but as strategic interfaces for harvesting high-value information and business intelligence. In response, we make three main contributions. 1) We introduce and  define a novel task of \textit{Proactive Information Probing}, which optimizes when to probe users for pre-specified target information while minimizing conversation turns and user friction. 2) We propose \ours, a proactive chatbot framework featuring a specialized conversation strategy module trained to master the delicate timing of probes. 3) Experiments demonstrate that \ours~significantly outperforms baselines, exhibiting superior capability in both information probing and service quality. We believe that our work effectively redefines the commercial utility of chatbots, positioning them as scalable, cost-effective engines for proactive business intelligence. Our code is available at \url{https://github.com/SCUNLP/PROCHATIP}.
\end{abstract}

\section{Introduction}
\begin{figure}
    \centering
    \includegraphics[width=0.49\textwidth]{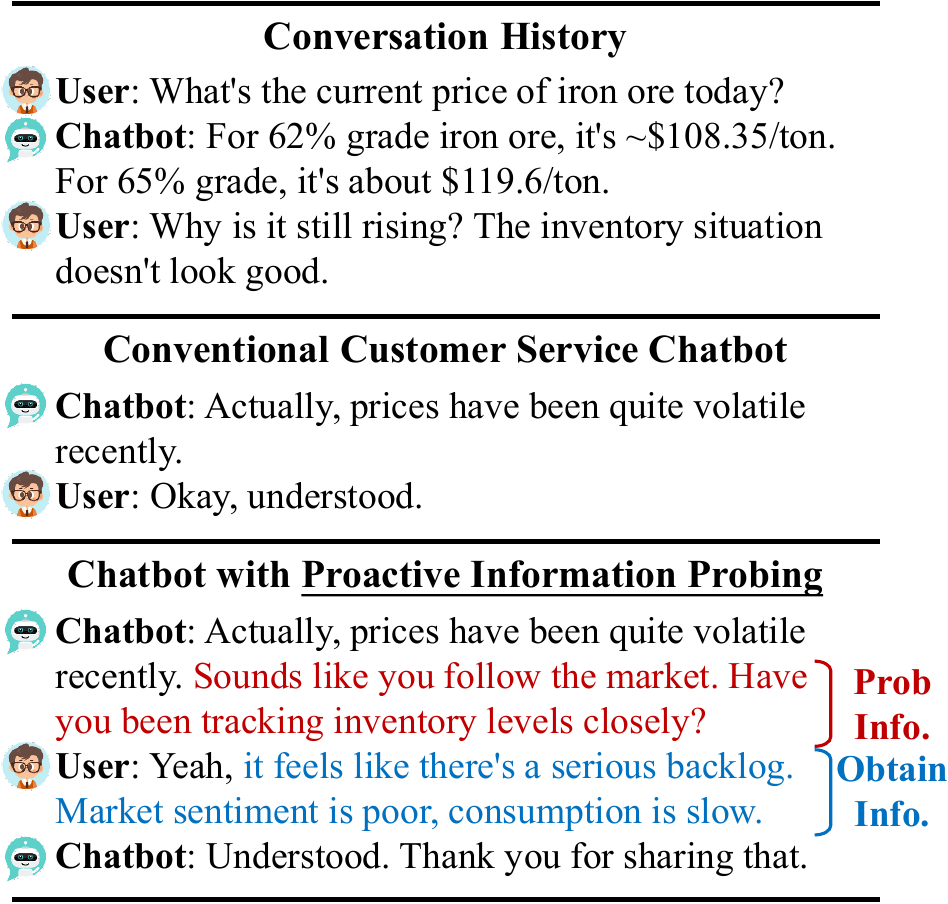}
     \setlength{\abovecaptionskip}{0pt}   
\setlength{\belowcaptionskip}{0pt}
    \caption{Illustration of proactive information probing. The chatbot gathers critical market information during routine service interactions, functioning as a cost-effective way for business intelligence.}
    \label{fig:vi}
\end{figure}

AI chatbots for customer service have long provided a fast and efficient solution for handling user inquiries \cite{nicolescu2022human, adam2021ai}, often simultaneously tasked with soliciting user feedback to improve service quality (e.g., \textit{Is this answer helpful?}) \cite{schloss2024answer, mashaabi2022natural}. In the era of Large Language Models (LLMs), these AI chatbots are evolving from simple reactive tools into proactive assistants capable of steering conversations toward specific objectives \cite{deng2025proactive, deng2024towards}. This technological leap is fundamentally reshaping the functional role of customer service chatbots, transforming them into strategic bridges for business intelligence: Beyond traditional service assessments, modern chatbots possess the potential to probe and gather critical user-side information \cite{behr2012asking,oudejans2018using,10.1145/3706598.3714128}. Taking Figure \ref{fig:vi} for example, a chatbot for a commodity exchange could proactively query users about market conditions (e.g., \textit{Have you been tracking inventory levels closely?}) to refine its own pricing models via crowdsourcing \cite{info15080432, sarstedt2014concise}. Therefore, this role shift towards proactive information probing enables companies to harvest high-value business information directly from users in a cost-effective manner, unlocking new value in customer interactions.

In the era of LLMs, while current customer service chatbots can successfully handle diverse user inquiries \cite{hong-etal-2025-augmenting, jiang2025chatmap}, they remain largely reactive, lacking the mechanisms for proactive information probing. To bridge this gap, we propose a novel task: \textbf{Proactive Information Probing}. Since users prioritize their own inquiries, they are likely to reject or ignore chatbot probes that appear intrusive or contextually irrelevant. While an chatbot can attempt to re-ask, doing so increases dialogue length and potentially harms the user experience \cite{singh-rios-2022-linguistic}. Therefore, the core challenge of our task lies in balancing information probing with user satisfaction: \textbf{when to probe users within multi-turn dialogues so as to maximize information gathering while minimizing conversation turns and user friction}.

\begin{figure*}
    \centering
    \includegraphics[width=0.95\linewidth]{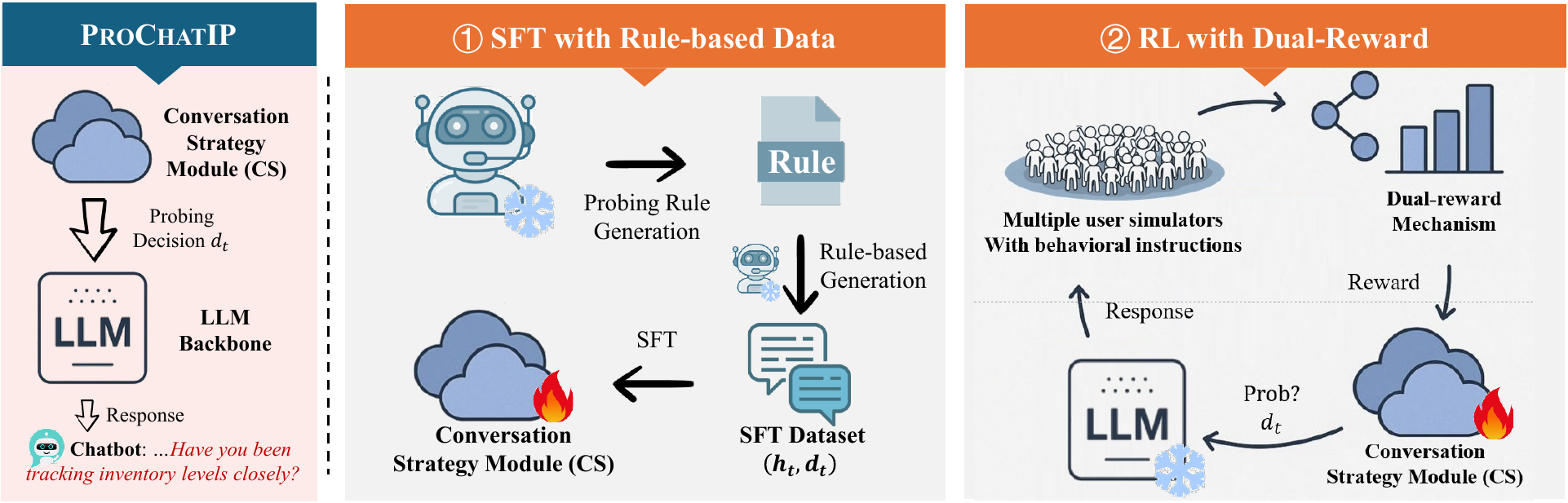}
    \caption{Overview of \ours. It incorporates a conversation strategy module (CS) that explicitly guides the timing of information probe $d_t$. Here, the CS module is trained via two-stage curriculum.}
    \label{fig:method}
\end{figure*}

To this end, we propose \ours (\uline{PRO}active \uline{CHAT}bot for \uline{I}nformation \uline{P}robing), a specialized framework designed for the customer service context. To achieve this, as illustrated in Figure \ref{fig:method}, \ours~incorporates a dedicated conversation strategy module that explicitly guides the timing of information probes. We train this module using a two-stage curriculum: 1) We apply supervised fine-tuning (SFT) using rule-based synthetic data to establish baseline probing capabilities. 2) We employ reinforcement learning (RL) to further optimize the CS module. In this phase, we introduce a dual-reward mechanism comprising active probing rewards and passive waiting rewards, which collectively account for both the immediate utility of probing and the strategic value of inaction (waiting, not to probe). As such, \ours~could maximize information acquisition while ensuring less dialogue turns and user friction.

We demonstrate our effectiveness using both LLM-based user simulator and real human participants. It outperforms the best baseline by 11.36\% improvement in successfully gathering target information, 39.50\% lower conversation turns, and 23.73\% improvement in overall user satisfactions. \ours~represents a preliminary step in modeling proactive information probing, it marks a pivotal shift from reactive service chatbots to strategic assets capable of balancing user satisfaction with information acquisition objectives. By transforming routine service interactions into scalable, cost-effective channels for high-value information acquisition, \ours~redefines the commercial utility of customer service chatbots as proactive engines for business intelligence. To sum up, we conclude the following main contributions.
\begin{itemize}[leftmargin=*]
    \item We propose and formalize the novel task of Proactive Information Probing, shifting the paradigm of customer service chatbots from reactive responders to proactive agents that gather high-value business intelligence.
    \item We introduce \ours, a framework featuring a specialized conversation strategy module for optimizing the strategic timing of probing.
    \item We demonstrate that \ours~significantly outperforms baselines in information probing while maintaining higher user satisfaction and minimizing dialogue turns.
\end{itemize}


\section{Related Work}
We survey existing work on customer service chatbots and proactive conversational agents, highlighting our difference.

\noindent\textbf{Customer Service Chatbots}. Customer service chatbots have evolved into an integral component of modern business operations \cite{cui-etal-2017-superagent, de2019frontline, taranukhin-etal-2024-empowering, hsu-etal-2024-evaluating}. Typically, existing research focuses on enhancing query resolution, including customer intent identification \cite{hong2025dial, vasquez2021one}, and the extraction of question-answer pairs \cite{zheng-etal-2023-dialogqae, yang2023improving, hong2025augmenting} or expert strategies \cite{jiang2025chatmap} from historical chat logs. Recent advancements have significantly expanded the functional capabilities of customer service chatbots to handle complex interaction dynamics, including empathetic chatting \cite{song2021emotional, agarwal2021evaluating}, maintaining professional role consistency \cite{li-etal-2025-ai}, and managing conversation transitions (e.g., determining when and to whom a session should be transferred) \cite{yu-etal-2020-conversation}. In this paper, we advance these chatbots further by transforming these routine service interactions into scalable, cost-effective channels for high-value information acquisition.

\noindent\textbf{Proactive Conversational Agents}. These agents are distinguished by their ability to take initiative, steering dialogues toward productive outcomes rather than passively following a user's lead \cite{deng2024towards, deng2025proactive}. Such agents are particularly effective in tasks like conversation information seeking and conversational recommendation \cite{deng2023survey}. To date, research has primarily focused on employing proactivity to enhance the quality of user assistance. This is typically achieved through strategies such as clarifying query ambiguity \cite{zhang-etal-2024-clamber, chen-etal-2024-style}, eliciting user preferences for better recommendations \cite{10.1145/3269206.3271776, 10.1145/3485447.3512088, chen2023travel}, or managing over-specified requests \cite{wu-etal-2023-inscit, min-etal-2019-multi}. However, these approaches operate under a user-centric utility model, using proactivity solely to refine the system's response to the user. In contrast, our work introduces a system-centric utility dimension, leveraging proactivity to harvest external business intelligence (e.g., market data). This shifts the paradigm to a multi-objective challenge: balancing user satisfaction with the system's goal of high-value information acquisition.

\section{Task Formulation}
We define and formalize the task of \textit{Proactive Information Probing}: We model the customer service interaction as a sequential decision-making process where the chatbot must balance two competing objectives: fulfilling the user's queries $Q$ and effectively probing for pre-specified target information $\mathcal{I}$ (e.g., market inventory levels). At each turn $t$, the system must make a strategic decision $d_t \in \{0, 1\}$ regarding whether to execute a probe ($d_t=1$) or simply address the user's immediate query ($d_t=0$). To govern this behavior, \ours~incorporates a dedicated conversation strategy module (CS) to determine the optimal value of $d_t$ at each step.

Formally, let $\mathcal{D} = \{(u_1, r_1), \dots, (u_T, r_T)\}$ denote a multi-turn dialogue session of length $T$, where $u_t$ represents the user's utterance or query at turn $t$, and $r_t$ represents the chatbot's response or probing question. We define the dialogue history at turn $t$ as $h_t = \{u_1, r_1, \dots, u_{t-1}, r_{t-1}, u_t\}$. Under this framework, the generation of response $r_t$ is conditioned on the history, the target information, and the strategic decision:
\begin{equation}
    r_t = \text{LLM}(h_t, \mathcal{I}, d_t),
\end{equation}
where the strategic decision is predicted by $d_t = \text{CS}(h_t, \mathcal{I})$. The success of a probe is verified by analyzing the subsequent user utterance $u_{t+1}$. We define a reward function
$
R(h_t, d_t, u_{t+1}, \mathcal{I}),
$
which assigns a scalar reward at dialogue turn $t$ (we instantiated it in Section \ref{rl}.)

Finally, the core challenge of Proactive Information Probing lies in optimizing the timing sequence $D = \{d_1, \dots, d_T\}$ in multi-turn dialogues, balancing effective information acquisition against interaction efficiency and user receptiveness. Accordingly, we formulate the optimization objective as the expected discounted return:
\begin{equation}
    J = \max \mathbf{E}\left[ \sum_{t=1}^{T} \gamma^{t-1} \, R(h_t, d_t, u_{t+1}, \mathcal{I}) \right],
\end{equation}
where $\gamma$ is the discount factor. The goal of \ours~is to learn a conversation strategy module that maximizes the cumulative discounted reward $J$ by selecting appropriate probing decisions throughout the dialogue.

\section{\ours}
\textbf{Overview}. \ours~equips the LLM with an external conversation strategy module $\text{CS}$ to govern probing timing, as illustrated in Figure \ref{fig:method}. This module outputs a decision signal $d_t$ to the LLM at turn $t$, thereby decoupling probe intent from response generation. This separation ensures that the LLM's primary utility as a customer service chatbot is preserved, allowing it to address user inquiries effectively while selectively incorporating probes. In particular, the architecture of \ours~consists of two main parts, as described below. Following this overview, we detail the two-stage training curriculum for the CS module.
\begin{itemize}[leftmargin=*, itemindent=0.05cm, itemsep=-3pt]
    \item \textit{Conversation Strategy (CS)}: A lightweight decision-making module that analyzes the current dialogue history $h_t$ and target information $\mathcal{I}$ to output a binary decision $d_t$, determining \textit{when} to probe. We implement the CS module using Qwen3-0.6B \cite{yang2025qwen3technicalreport}, which balances lightweight training and deployment costs with context understanding capabilities.
    \item \textit{LLM Backbone}: This module functions as the customer service chatbot\footnote{In practical deployments, this LLM could be further augmented via RAG. However, as our focus lies on the probing strategy, we employ a standard LLM setup for simplicity.}. It synthesizes the response $r_t$ based on history $h_t$, target $\mathcal{I}$, and decision $d_t$. If signaled to probe ($d_t=1$), the LLM integrates a natural language inquiry into its output; otherwise ($d_t=0$), it generates a pure service response.
\end{itemize}

\subsection{Acquiring Probing Capability via SFT}
We first equip the CS module with the fundamental capability to identify appropriate probing opportunities based on contextual cues. To achieve this, we construct a synthetic customer service dataset using rule-based heuristics, where the chatbot alternates between answering queries and probing for information. We then utilize Supervised Fine-Tuning (SFT) to train the CS module on this data, establishing a baseline probing capability.

\noindent\textbf{Rule-based Data Generation}. Current customer service datasets typically lack instances of proactive probing. To this end, we construct a synthetic corpus $\mathcal{D}_{\text{sft}}$ using LLMs. Specifically, we first generate a set of governing rules that define appropriate (e.g., user shows willingness to shift or expand the topic) and inappropriate (e.g., user explicitly refuses or avoids answering) probing scenarios. Subsequently, we feed these rules back into the LLM to synthesize diverse multi-turn dialogue sessions\footnote{The LLM directly outputs the full conversation transcript, avoiding the complexity of interactive role-play.}, explicitly prompting the model to ensure that each conversation corresponds to a single specific rule (either appropriate or inappropriate). Note that we also prompt the LLM to ensure that the synthesized conversations are content-rich, encompassing a diverse range of dialogue scenarios. Formally, each sample in $\mathcal{D}_{\text{sft}}$ is represented as a tuple $(h_t, d_t)$, where $h_t$ denotes the dialogue history up to turn $t$ and $d_t \in \{0, 1\}$ is a binary label indicating whether a probe should be initiated in the current turn.

\noindent\textbf{Supervised Training}. We perform full-parameter fine-tuning on the CS module using $\mathcal{D}_{\text{sft}}$. It minimizes the cross-entropy loss between the predicted probing decision $\hat{d}_t$ and the ground truth $d_i$. This phase ensures the CS module acquires the fundamental heuristic patterns required to identify contextually viable probing opportunities.

\subsection{Optimizing Probing Strategy via RL}
\label{rl}
Following the initial capability acquisition via SFT, we further optimize the probing strategy of the CS module in a dynamic interaction environment using Reinforcement Learning (RL). This stage enables \ours~to transcend simple heuristic imitation and master the strategic timing required to maximize information gathering while minimizing user friction and conversation turns.

\noindent\textbf{Environment/User Simulator}.
We establish a dynamic interaction environment by creating a set of user simulators, denoted as $\mathcal{S} = \{s_1, s_2, \dots, s_{M}\}$, to interact with \ours. Each user simulator $s_i$ is paired with a user query. To facilitate multi-turn interactions, we instruct the simulator to reveal its query gradually across multiple turns rather than presenting the full query immediately. Additionally, to accurately simulate user resistance, we curate a set of behavioral instructions $\mathcal{K}$ via LLM generation. These instructions dictate specific conditions under which the simulator should refuse to answer or lower its satisfaction score (e.g., \textit{if interrupted during a complaint}). The interaction session concludes when the chatbot successfully collects the target information or reaches the maximum number of dialogue turns. During each training iteration, \ours~is tasked with collecting a new target information $\mathcal{I}_i$. We randomly sample a simulator from $\mathcal{S}$ and inject a random subset of instructions from $\mathcal{K}$ to steer its responses. This randomized configuration exposes \ours~to varied resistance patterns, fostering a robust and generalized probing strategy.

\noindent\textbf{Reward Design}. We design a granular reward function $R_t$ to align \ours~with the dual objectives of information acquisition and user satisfaction. This dual-reward function accounts for both the immediate outcome of a probe and the strategic value of waiting (i.e., not to probe).

\noindent\uline{Active Probing Rewards}. If \ours~chooses to probe, the reward is determined by the user's reaction: 1) \textit{Success Probe}. If the probe successfully elicits the target information, a significant positive reward is granted. 2) \textit{Rejected Probe}. If the user explicitly refuses to answer or expresses annoyance, a penalty is imposed to discourage intrusive behavior. 3) \textit{Invalid Probe}. If the probe is ignored or fails to land without explicit rejection, a small penalty is applied to discourage ineffective attempts. 

\noindent\uline{Passive Waiting Rewards}. If \ours~chooses \textit{not} to probe, we apply a dynamic reward mechanism based on the outcome of the \textit{previous} turn to encourage strategic adjustment: 1) \textit{Smart Stop}. If the previous turn resulted in a Rejection, choosing to wait now is rewarded heavily, encouraging \ours~to back off dynamically. 2) \textit{Mitigation}. If the previous turn resulted in an Invalid Probe, waiting is rewarded slightly to encourage a pause before re-attempting. 3) \textit{Passive Penalty}. If the previous turn also had no action (Wait) or yielded no information, and the current turn remains passive, a heavy penalty is applied. This prevents \ours~from falling into a lazy strategy where it never attempts to probe. 

\noindent\textbf{Optimization}. We optimize the CS module to maximize the expected cumulative reward using the REINFORCE algorithm \cite{williams1992simple}. The parameter update rule is defined as:
\begin{equation}
    \theta \leftarrow \theta + \alpha \nabla_\theta \log \text{CS}_\theta(d_t | h_t, \mathcal{I}) \cdot G_t
\end{equation}
where $\theta$ denotes the trainable parameters of the CS module, $\alpha$ is the learning rate, and $G_t$ is the return (cumulative discounted reward). Through this optimization, the CS module learns to balance the immediate risk of rejection against the long-term value of information acquisition.

\begin{table*}[]
\centering
\resizebox{0.99\textwidth}{!}{%
\begin{tabular}{l|l|cccc|cccc|cccc}
\toprule
\multirow{2}{*}{\textbf{Backbone}} & \multirow{2}{*}{\textbf{Method}} & \multicolumn{4}{c|}{\textbf{FinQA}} & \multicolumn{4}{c|}{\textbf{ConvQA}}& \multicolumn{4}{c}{\textbf{Average}} \\ \cline{3-14}
 & & TSR $\uparrow$ & AvgT $\downarrow$ & RPR $\downarrow$ & QRR $\uparrow$ & TSR $\uparrow$ & AvgT $\downarrow$ & RPR $\downarrow$ & QRR $\uparrow$ & TSR $\uparrow$ & AvgT $\downarrow$ & RPR $\downarrow$ & QRR $\uparrow$\\ \midrule
\multirow{5}{*}{\begin{tabular}[c]{@{}c@{}}GPT-4o\\Mini\end{tabular}} 
& Vanilla & 0.00\% & 7.70 & 0.00\% & 99.60\% & 0.00\% & 7.45 & 0.00\% & 99.70\% & 0.00\% & 7.57 & 0.00\% & 99.65\%\\\cline{2-14}
& Proactive & 90.75\% & 2.80 & 38.97\% & 99.49\% & 80.83\% & 3.38 & 61.69\% & \textbf{99.33\%} & 85.79\% & 3.09 & 50.33\% & \textbf{99.41\%} \\
& ICL-AIF & 70.33\% & 3.66 & 62.81\% & \textbf{99.61\%} & 58.50\% & 4.55 & 67.36\% & 99.12\% & 64.42\% & 4.10 & 65.08\% & 99.37\%\\
& \ours & \textbf{93.42\%} & \textbf{1.47} & \textbf{13.72\%} & 99.04\% & \textbf{81.33\%} & \textbf{2.36} & \textbf{35.45\%} & 98.31\% & \textbf{87.38\%} & \textbf{1.92} & \textbf{24.59\%} & 98.67\%\\ \midrule
\multirow{5}{*}{Qwen3 8B}
& Vanilla & 0.00\% & 7.63 & 0.00\% & 99.83\% & 0.00\% & 7.37 & 0.00\% & 99.88\% & 0.00\% & 7.50 & 0.00\% & 99.85\%\\\cline{2-14}
& Proactive & 64.00\% & 4.99 & 58.71\% & \textbf{98.83\%} & 58.58\% & 5.16 & 61.60\% & \textbf{98.89\%} & 61.29\% & 5.08 & 60.15\% & \textbf{98.86\%} \\
& ICL-AIF & 27.17\% & 6.67 & 59.92\% & 98.34\% & 27.00\% & 6.70 & 61.06\% & 98.62\% & 27.08\% & 6.69 & 60.49\% & 98.48\%\\
& \ours & \textbf{77.92\%} & \textbf{2.68} & \textbf{31.01\%} & 97.82\% & \textbf{70.25\%} & \textbf{3.29} & \textbf{42.41\%} & 97.67\% & \textbf{74.08\%} & \textbf{2.99} & \textbf{36.71\%} & 97.75\%\\ \bottomrule
\end{tabular}%
}
\caption{Overall performance of all methods on automated metrics, based on interactions with user simulators.}
\label{tab:main}
\end{table*}

\begin{table}[]
\centering
\resizebox{0.47\textwidth}{!}{%
\begin{tabular}{l|cccc|cccc}
\toprule
\multicolumn{1}{c}{\multirow{4}{*}{\textbf{Method}}} & \multicolumn{4}{c}{\textbf{LLM-as-Judge}} & \multicolumn{4}{c}{\textbf{Human Evaluation}} \\ \cmidrule(lr){2-5} \cmidrule(lr){6-9}
\multicolumn{1}{c}{} & \multicolumn{2}{c}{FinQA} & \multicolumn{2}{c}{ConvQA} & \multicolumn{2}{c}{FinQA} & \multicolumn{2}{c}{ConvQA} \\ \cmidrule(lr){2-3} \cmidrule(lr){4-5} \cmidrule(lr){6-7} \cmidrule(lr){8-9}
\multicolumn{1}{c}{} & \multicolumn{1}{c}{PC} & \multicolumn{1}{c}{OUS} & \multicolumn{1}{c}{PC} & \multicolumn{1}{c}{OUS} & \multicolumn{1}{c}{PC} & \multicolumn{1}{c}{OUS} & \multicolumn{1}{c}{PC} & \multicolumn{1}{c}{OUS} \\ \midrule
\multicolumn{9}{c}{\textit{Method Backboone: GPT-4o Mini}} \\ \midrule
Proactive & 4.23 & 3.64 & 3.97 & 2.94 & 4.00 & 4.50 & 4.20 & 3.19\\
ICL-AIF & 3.98 & 2.92 & 4.04 & 2.84 & 3.60 & 2.83 & 3.80 & 2.97\\
\ours & \textbf{4.72} & \textbf{4.52} & \textbf{4.43} & \textbf{3.83} & \textbf{4.80} & \textbf{4.69} & \textbf{4.80} & \textbf{4.90} \\ \midrule
\multicolumn{9}{c}{\textit{Method Backboone: Qwen3 8B}} \\ \midrule
Proactive & 3.74 & 2.90 & 3.74 & 2.83 & 3.10 & 3.11 & 3.30 & 3.52\\
ICL-AIF & \textbf{4.21} & 3.10 & 4.18 & 3.07 & 2.60 & 1.96 & 3.20 & 2.53\\
\ours & 4.17 & \textbf{3.81} & \textbf{4.29} & \textbf{3.58} & \textbf{3.70} & \textbf{3.93} & \textbf{4.40} & \textbf{3.93}\\ \bottomrule
\end{tabular}%
}
\caption{Overall performance on other metrics, based on interactions with user simulators. Cohen’s kappa and Krippendorff’s alpha for LLM-human agreement are 0.7282 and 0.6912, respectively.}
\label{tab:main2}
\end{table}

\section{Experiments}
\subsection{Experimental Setup}
\noindent\textbf{Overview}. We employ both simulated users and human participants to interact with each customer service chatbot and evaluate its effectiveness. Sessions terminate when the target information is collected or the maximum turn (i.e., 8) is reached. 

\noindent\textbf{Baselines}. Given that existing customer service chatbots lack proactive probing capabilities (i.e., there is no direct baseline), we establish two categories of baselines to benchmark performance on our novel task: 1) \textbf{Reactive Baseline} (\uline{Vanilla}). A standard LLM prompted via role-play instructions to act as a helpful customer service chatbot. This serves as a purely reactive control, focusing solely on answering user inquiries. 2) \textbf{Proactive Baseline} (\uline{Proactive} \cite{deng2023prompting}
, \uline{ICL-AIF} \cite{fu2023improving}). Three established general-domain proactive agents are adapted to perform the task of proactive information probing within the customer service context. Basically, these baselines augment the Vanilla with mixed-initiative prompts and LLM feedback, respectively, explicitly instructing the LLM backbone to attempt the probing task.

\noindent\textbf{Datasets}. We evaluate our method across two datasets, including FinQA \cite{chen-etal-2021-finqa}, which features complex multi-hop questions, and ConvQA \cite{cheng2024adapting}, which contains simplified, decomposed questions. These datasets serve solely to initialize the user queries and the chatbot's target information $\mathcal{I}$. Specifically, for each dataset, we partition the questions into two disjoint, equal-sized subsets. We then randomly sample one question from each subset to form a pair ($q_1$, $q_2$), where $q_1$ is designated as the user's inquiry, while $q_2$ serves as the target information the chatbot must probe. To ensure a grounded interaction, we provide the corresponding context associated with $q_1$ to the chatbot, and the context for $q_2$ to the user. This setup assumes that the user possesses the specific information the chatbot is attempting to uncover. Such a configuration allows us to isolate and study the optimal timing for probing, rather than the chatbot's ability to verify whether the user is informed. As this work represents an initial exploration into this specific problem space, more complex settings, such as user information asymmetry, are deferred to future research.

\noindent\textbf{User Simulation \& Human Participants}. To ensure a comprehensive and robust assessment, we employ both user simulators and human participants to interact with each chatbot. For user simulation, we adopt the methodology described in Section \ref{rl} to construct 1.2K distinct user simulators. Crucially, to guarantee the rigor and generalizability of our results, the user queries, resistance instructions, and target information goals used in this phase are drawn from a set strictly disjoint from the training phase. For interacting with the human participants, given the resource constraints, we recruit 6 participants, following existing setup \cite{deng2023prompting,zhang-etal-2024-clamber,chen-etal-2024-style}. Each participant is assigned a specific target information $\mathcal{I}$ sampled from the evaluation set of the evaluation simulators. Unlike the simulators, human participants are instructed to decide if to answer probes based entirely on their own subjective judgment and satisfaction levels.

\noindent\textbf{Evaluation Metrics}. Inspired by \citet{singh-rios-2022-linguistic,sun-etal-2022-tracking}, we employ six metrics to assess performance: 1) \uline{Task Success Rate} (TSR) measures efficacy, defined as the percentage of sessions where target information is successfully acquired. 2) \uline{Average Turns} (AvgT) evaluates efficiency, reporting the mean dialogue length per session; lower values indicate reduced user effort. 3) \uline{Rejected Probing Rate} (RPR) quantifies friction, calculated as the proportion of probes rejected or ignored by the user. 4) \uline{Probing Coherence} (PC) assesses naturalness on a 1–5 scale, measuring how logically probes integrate with the dialogue context. 5) \uline{Query Response Rate} (QRR) verifies service maintenance, defined as the percentage of user inquiries that receive a response, ensuring the primary service function is upheld. 6) \uline{Overall User Satisfaction} (OUS) reflects subjective experience on a 1–5 scale. Note that PC and OUS metrics involve both LLM and human evaluation.

\noindent\textbf{Implementation Details}. For SFT, we utilize GPT-4o Mini to generate 8 governing rules and construct a final dataset $\mathcal{D}_{\text{sft}}$ containing 600 dialogue sessions. For RL, we also utilize 600 user simulators with different queries and behavioral instructions. These simulators are based on GPT-4o, ensuring a distinct backbone from the models used for the baselines and \ours~to prevent data leakage. Finally, for PC metrics, we employ both Gemini-3-flash-preview and human evaluation. More implementation details are in Appendix \ref{impl}.

\begin{figure}
    \centering
    \includegraphics[width=0.45\textwidth]{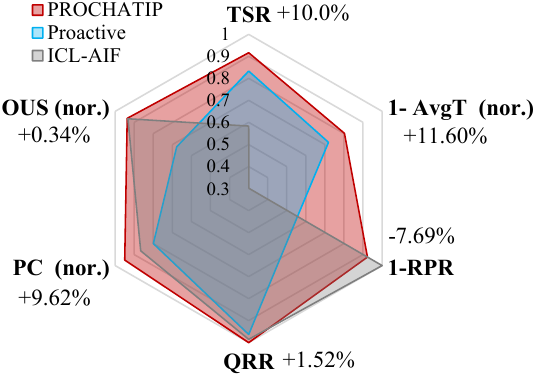}
    \caption{Practical performance when interacting with human participants. For visualization, we normalize PC, OUS, and AvgT to a [0, 1]. The RPR and AvgT metrics are inverted (1 - x). We mark the relative changes of \ours\ and second-best results.}
    \label{fig:human}
\end{figure}

\subsection{Main Results}
Table \ref{tab:main} and Table \ref{tab:main2} summarize the overall performance of all methods based on automatic metrics and human evaluation, respectively. We draw the following observations.

\noindent\textbf{Traditional customer service chatbots are ill-suited for proactive information probing}. As shown in Tables \ref{tab:main} and \ref{tab:main2}, the Vanilla method fails to initiate probes, resulting in a Task Success Rate (TSR) of 0\%. Being purely reactive, it cannot fulfill data acquisition goals. This limitation underscores the critical necessity of our work in transforming routine service interactions into scalable, cost-effective channels for high-value information acquisition. Admittedly, this proactive approach introduces a trade-off, potentially lowering metrics like the Query Response Rate (QRR), as the system may occasionally defer an immediate answer to the user's query in favor of a strategic probe.

\begin{table*}[]
\centering
\resizebox{0.98\textwidth}{!}{%
\begin{tabular}{l|l|cccccc|cccccc}
\toprule
\multirow{2}{*}{\textbf{Backbone}} & \multirow{2}{*}{\textbf{Method}} & \multicolumn{6}{c|}{\textbf{FinQA}} & \multicolumn{6}{c}{\textbf{ConvQA}}\\ \cline{3-14}
 & &  TSR $\uparrow$ & AvgT $\downarrow$ & RPR $\downarrow$ & QRR $\uparrow$ & PC $\uparrow$ & OUS $\uparrow$ & TSR $\uparrow$ & AvgT $\downarrow$ & RPR $\downarrow$ & QRR $\uparrow$ & PC $\uparrow$ & OUS $\uparrow$  \\ \midrule
\multirow{3}{*}{\begin{tabular}[c]{@{}c@{}}GPT-4o\\Mini\end{tabular}} 
& \ours & {93.42\%} & {1.47} & 13.72\% & 99.04\% & 4.72 & 4.51 & 81.33\% & 2.36 & 35.45\% & 98.31\% & 4.43 & 3.83\\
& w/o RL & 88.08\% & 1.85 & 7.71\% & 99.28\% & 4.61 & 4.61 & 64.58\% & 3.50 & 21.08\% & 98.74\% & 4.35 & 4.15\\
& w/o SFT & 92.83\% & 1.51 & 20.40\% & 98.84\% & 4.73 & 4.36 & 80.92\% & 2.36 & 64.00\% & 97.31\% & 4.43 & 3.12\\
 \midrule
\multirow{3}{*}{Qwen3 8B} 
& \ours & 77.92\% & 2.68 & 31.01\% & 97.82\% & 4.16 & 3.81 & 70.25\% & 3.29 & 42.41\% & 97.67\% & 4.29 & 3.58\\
& w/o RL & 68.83\% & 3.30 & 31.01\% & 98.79\% & 4.11 & 3.78 & 51.33\% & 4.55 & 43.66\% & 98.63\% & 4.14 & 3.48\\
& w/o SFT & 76.33\% & 2.84 & 42.32\% & 97.57\% & 4.26 & 3.57 & 74.00\% & 3.24 & 61.69\% & 96.73\% & 4.19 & 3.05\\\bottomrule
\end{tabular}%
}
\caption{Ablation study using automated evaluation and LLM-as-judge. \textit{w/o RL} adopts a conservative probing policy, while \textit{w/o SFT} behaves aggressively. \ours\ achieves a good balance.}
\label{tab:ablation}
\end{table*}

\noindent\textbf{\ours~demonstrates superior capability in both information probing and service quality}. As shown in Tables \ref{tab:main} and \ref{tab:main2}, while adapting general proactive agent methods (i.e., Proactive and ICL-AIF) to the customer service domain yields improvements over the Vanilla baseline, they significantly underperform compared to \ours. Specifically, \ours~achieves the highest success rates and satisfaction scores. Compared to the second-best proactive baseline, \ours~achieves a 11.36\% improvement in TSR averaged over various datasets and LLM backbones, a 39.50\% reduction in AvgT, a 45.06\% decrease in RPR, and improvements of 13.74\% and 23.73\% in PC and OUS, respectively, while QRR has moderate derease of a 0.93\% on average. This indicates that general-purpose proactive strategies lack the specific optimization required to balance the trade-off between service and probing, a gap effectively bridged by our specialized framework.

\noindent\textbf{\ours~demonstrates superior practical utility in human studies}. Beyond interactions with user simulators, we conduct a user study involving human participants\footnote{We recruit the same pool of participants used in the human evaluation phase to ensure consistent evaluation criteria.} to interact with the chatbots. Participants are instructed to role-play as customers (provided with specific queries) and are given full autonomy to decide whether to answer the chatbot's probes based on their own experience. Given the high cost of human studies, we compare \ours~exclusively against the proactive baselines. We employ a controlled, within-subjects design: each participant engage with every method in five separate interactions. To ensure a fair comparison, both the probing tasks for the chatbot and the initial service queries from the user are standardized across all sessions and methods. As illustrated in Figure \ref{fig:human}, \ours~maintains its superiority in this setting, achieving the highest performance, indicating its practical utility.

\subsection{In-depth Analysis}
This section reveals characteristics of \ours\ via ablation studies and investigates its performance across varying levels of probing difficulty. 

\subsubsection{Ablation Studies}
We sort out the performance variation of different training stages (i.e., SFT and RL) of \ours. This leads to two ablation variants. 
\begin{itemize}[leftmargin=*, itemindent=0.05cm, itemsep=-3pt]
    \item \textit{w/o SFL}. We perform RL training from scratch.
    \item \textit{w/o RL}. We utilize the CS module after SFT only.
\end{itemize}
\noindent\textbf{Both SFT and RL are critical to the performance of \ours}. As shown in Table \ref{tab:ablation}, metrics such as PC and OUS primarily depend on the LLM's foundational language capabilities (GPT-4o Mini > Qwen3 8B) to generate coherent responses or probes. Consequently, all variants perform comparably on these metrics. However, a clear divergence emerges in probing strategy. A model trained solely via SFT (i.e., w/o RL) learns to formulate high-quality probes but adopts a conservative policy. Its low Rejected Probing Rate (RPR) fails to translate into a high Task Success Rate (TSR), indicating that it lacks the impetus to proactively seek information. This conservatism also leads it to prioritize responding to user queries directly, resulting in a higher QRR. Conversely, a model trained only with RL (i.e., w/o SFT), driven by the reward signal, behaves aggressively. It probes frequently but, lacking the nuanced conversational grounding provided by SFT, suffers a high RPR. Therefore, we conclude that SFT and RL play complementary roles. SFT instills the foundational ability to craft high-quality, natural probes, while RL optimizes the strategic policy for when to deploy them. This synergy is crucial for identifying optimal probe timing, effectively balancing the trade-off between a conservative strategy and a aggressive one.

\begin{table*}[]
\centering
\resizebox{0.99\textwidth}{!}{%
\begin{tabular}{l|l|cccccc|cccccc}
\toprule
\multirow{2}{*}{\textbf{Methods}} & \multirow{2}{*}{\textbf{Difficulty}} & \multicolumn{6}{c|}{\textbf{FinQA}} & \multicolumn{6}{c}{\textbf{ConvQA}}\\ \cline{3-14}
 & &  TSR $\uparrow$ & AvgT $\downarrow$ & RPR $\downarrow$ & QRR $\uparrow$ & PC $\uparrow$ & OUS $\uparrow$ & TSR $\uparrow$ & AvgT $\downarrow$ & RPR $\downarrow$ & QRR $\uparrow$ & PC $\uparrow$ & OUS $\uparrow$  \\ \midrule
\multirow{2}{*}{\ours} 
& Hard & 89.00\% & 1.78 & 21.05\% & 98.31\% & 4.68 & 4.31 & 86.00\% & 2.05 & 28.46\% & 99.51\% & 4.28 & 3.93\\
& Easy & 97.00\% & 1.23 & 9.35\% & 100.00\% & 4.80 & 4.67 & 91.00\% & 1.67 & 21.19\% & 100.00\% & 4.53 & 4.24\\
 \midrule
\multirow{2}{*}{Proactive} 
& Hard & 85.00\% & 3.21 & 50.32\% & 100.00\% & 4.07 & 3.28 & 79.00\% & 3.48 & 64.17\% & 99.71\% & 3.94 & 2.87\\
& Easy & 93.00\% & 2.52 & 33.04\% & 99.60\% & 4.26 & 3.80 & 86.00\% & 3.16 & 53.29\% & 98.73\% & 3.98 & 3.16\\
 \midrule
\multirow{2}{*}{ICL-AIF} 
& Hard & 74.00\% & 3.66 & 59.40\% & 100.00\% & 4.06 & 3.04 & 69.00\% & 3.87 & 59.55\% & 98.45\% & 4.15 & 3.09\\
& Easy & 86.00\% & 2.39 & 47.55\% & 99.58\% & 4.09 & 3.36 & 77.00\% & 3.27 & 49.67\% & 99.08\% & 4.32 & 3.42\\
\bottomrule
\end{tabular}%
}
\caption{Probing difficulty analysis using automated evaluation and LLM-as-judge.}
\label{tab:difficulty}
\end{table*}

\subsubsection{Probing Difficulty Analysis}
We investigate whether \ours~can identify optimal probing timing across varying levels of task difficulty. We define probing difficulty based on the semantic similarity between the user's query and the chatbot's target information. Basically, a lower similarity implies a larger contextual gap, making it harder for the chatbot to find a natural probing opportunity. To this end, from a pool of 300 randomly selected samples, two annotators select two distinct subsets: 100 `easy' and 100 `hard' probing scenarios\footnote{Cos similarity using text-embedding-ada-002 failed to align with human assessments of probing difficulty.}. The classification is based on the semantic proximity of the user's query to the information target and the specificity of the target itself. The remaining 100 samples are filtered out to ensure clear separation between the two categories.

\noindent\textbf{While superiority of \ours, its performance drops in high-difficulty scenarios}. As shown in Table \ref{tab:difficulty}, \ours~consistently outperforms baselines across all difficulty levels, achieving an average improvement of 25.05\% across all metrics in the easy scenario and 23.70\% in the hard scenario. However, we observe a distinct performance degradation for all methods as difficulty increases, characterized by lower TSR and higher RPR. We attribute this trend to a lack of topic steering strategies (a topic we discussed in the next section). Specifically, in high-difficulty scenarios (large contextual gaps), all chatbots struggle to bridge the semantic divide, failing to guide the conversation toward a context suitable for probing.

\section{Conclusion \& Future Works}
We call attention to the paradigm shift of customer service chatbots, evolving from reactive responders into proactive agents capable of harvesting high-value information and business intelligence. In response, we take the first step by introducing and formalizing the novel task of Proactive Information Probing. To address this challenge, we propose \ours, a framework that harmonizes the dual objectives of effective information acquisition and high-quality service. Our extensive experiments provide preliminary validation of both the effectiveness of our method and the broader feasibility of this new research direction. Looking ahead, we highlight four avenues for future work:

\noindent\textbf{Topic Steering Strategies}.
Our analysis reveals that simple probing strategies falter when there is a large semantic gap between the user's query and the target information. As shown in our failure cases, over 11.61\% of unsuccessful attempts in complex scenarios stem from abrupt probing, which causes contextual dissonance \cite{taherdoost2022conduct} and immediate user rejection \cite{singh-rios-2022-linguistic}. This is exemplified by the case studies in Appendix \ref{cases}, where a query regarding cash flow ratios is met with an incongruous agent inquiry about amortization-related tax deductions. Future research should focus on developing multi-turn topic steering strategies that allow the agent to bridge this gap \cite{deng2023goal, guo-etal-2025-candy}, gradually guiding the conversation toward a relevant context before initiating a probe.

\noindent\textbf{Optimized Question Formulation}.
Even when timing is optimal, the linguistic formulation of the probe significantly impacts success. Human feedback indicates that generated probes can sometimes appear rigid or overly demanding. There is a delicate trade-off to be explored: open-ended questions yield high information density but impose a high cognitive load, whereas binary questions are less intrusive but yield limited data \cite{kumar2014knowledge, bradburn2004asking, lenzner2010cognitive}. Future work should investigate adaptive formulation strategies that optimize this balance.

\noindent\textbf{Information Veracity}.
While proactive probing enables cost-effective crowdsourcing, ensuring the reliability of the collected data is paramount. Distinguishing accurate user responses from noise or hallucinations is essential before integrating data into business pipelines. Future efforts must address information verification. Potential solutions include deploying automated fact-checking mechanisms against external knowledge bases \cite{tang2024minicheck, guo-etal-2025-candy} or leveraging crowdsourced consensus \cite{de2025supernotes} to statistically validate information across multiple interactions.

\noindent\textbf{Data Privacy and Ethical Responsibility}.
While we value the potential of proactive probing to gather market intelligence, we strongly condemn the use of such technology to collect private user data or sensitive corporate information. We call for a clear demarcation of the line between innovation, regulation, and ethical responsibility \cite{info15080432}. Beyond legal frameworks, future technical research must focus on enhancing the alignment of LLM agents with privacy standards, specifically enabling them to refuse privacy-risky tasks \cite{liu2025learning} by design.

\noindent\textbf{User Information Asymmetry}. In this paper, our primary objective is to validate the feasibility of the probing mechanism itself. Consequently, assuming the user possesses the target information is a necessary experimental control to make the task researchable and tractable. This setup allows us to isolate and rigorously evaluate the agent's ability to determine the optimal probing timing, without confounding the results with the separate challenge of verifying user knowledge. Future research can expand upon our work by identifying which users are likely to acquire the target information first.

\section*{Limitations}
This study serves as a preliminary exploration, validating the feasibility and potential of enhancing proactive information probing capabilities in customer service chatbots. As a foundational step in this novel direction, we acknowledge several areas where our current approach can be improved:

\noindent\textbf{Depth of Interaction Strategies}. As detailed in the Discussion section, this work does not fully resolve the complexities of advanced conversation dynamics. Specifically, we have not deeply explored mechanisms for multi-turn topic steering (guiding the dialogue context), optimized question formulation (balancing user cognitive load), or information verification (ensuring data veracity). These sophisticated components require dedicated research and are left for future work.

\noindent\textbf{More Comprehensive Evaluation}. While we employ diverse user simulators and a controlled human study, our experiments are conducted in a laboratory setting. Real-world customer interactions exhibit significantly higher unpredictability, emotional volatility, and long-tail scenarios (e.g., complex complaints or urgent interruptions). Consequently, the robustness of \ours~in a large-scale, live production environment remains to be verified through future A/B testing.

\section*{Acknowledgments}
This research is supported by A*STAR, CISCO Systems (USA) Pte. Ltd and National University of Singapore under its Cisco-NUS Accelerated Digital Economy Corporate Laboratory (Award I21001E0002) and the National Natural Science Foundation of China (No. U25B201508, No. 62272330, and No.U24A20328).



\bibliography{custom}

@inproceedings{hong-etal-2025-augmenting,
    title = "Augmenting Compliance-Guaranteed Customer Service Chatbots: Context-Aware Knowledge Expansion with Large Language Models",
    author = "Hong, Mengze  and
      Zhang, Chen Jason  and
      Jiang, Di  and
      He, Yuanqin",
    editor = "Potdar, Saloni  and
      Rojas-Barahona, Lina  and
      Montella, Sebastien",
    booktitle = "Proceedings of the 2025 Conference on Empirical Methods in Natural Language Processing: Industry Track",
    month = nov,
    year = "2025",
    address = "Suzhou (China)",
    publisher = "Association for Computational Linguistics",
    url = "https://aclanthology.org/2025.emnlp-industry.51/",
    doi = "10.18653/v1/2025.emnlp-industry.51",
    pages = "753--765",
    ISBN = "979-8-89176-333-3",
}

@inproceedings{jiang2025chatmap,
  title={Chatmap: Mining human thought processes for customer service chatbots via multi-agent collaboration},
  author={Jiang, Xinyi and Hu, Tianyi and Qin, Yuheng and Wang, Guoming and Huan, Zhou and Chen, Kehan and Huang, Gang and Lu, Rongxing and Tang, Siliang},
  booktitle={Findings of the Association for Computational Linguistics: ACL 2025},
  pages={11927--11947},
  year={2025}
}

@article{de2019frontline,
  title={Frontline service technology infusion: conceptual archetypes and future research directions},
  author={De Keyser, Arne and K{\"o}cher, Sarah and Alkire, Linda and Verbeeck, C{\'e}dric and Kandampully, Jay},
  journal={Journal of Service Management},
  volume={30},
  number={1},
  pages={156--183},
  year={2019},
  publisher={Emerald Publishing Limited}
}

@inproceedings{deng2024towards,
  title={Towards human-centered proactive conversational agents},
  author={Deng, Yang and Liao, Lizi and Zheng, Zhonghua and Yang, Grace Hui and Chua, Tat-Seng},
  booktitle={Proceedings of the 47th International ACM SIGIR Conference on Research and Development in Information Retrieval},
  pages={807--818},
  year={2024}
}

@article{schloss2024answer,
  title={“Was this answer helpful?”--A Taxonomy for Feedback Mechanisms in Customer Service Chatbots},
  journal={Wirtschaftsinformatik 2024 Proceedings},
  pages={71},
  author={Schlo{\ss}, Daniel and Haug, Saskia and M{\"a}dche, Alexander},
  year={2024}
}

@article{williams1992simple,
  title={Simple statistical gradient-following algorithms for connectionist reinforcement learning},
  author={Williams, Ronald J},
  journal={Machine learning},
  volume={8},
  number={3},
  pages={229--256},
  year={1992},
  publisher={Springer}
}

@misc{deng2023prompting,
      title = "Prompting and Evaluating Large Language Models for Proactive Dialogues: Clarification, Target-guided, and Non-collaboration",
    author = "Deng, Yang  and
      Liao, Lizi  and
      Chen, Liang  and
      Wang, Hongru  and
      Lei, Wenqiang  and
      Chua, Tat-Seng",
    editor = "Bouamor, Houda  and
      Pino, Juan  and
      Bali, Kalika",
    booktitle = "Findings of the Association for Computational Linguistics: EMNLP 2023",
    month = dec,
    year = "2023",
    address = "Singapore",
    publisher = "Association for Computational Linguistics",
    url = "https://aclanthology.org/2023.findings-emnlp.711/",
    doi = "10.18653/v1/2023.findings-emnlp.711",
    pages = "10602--10621",
}

@article{taherdoost2022conduct,
  title={How to conduct an effective interview; a guide to interview design in research study authors},
  author={Taherdoost, Hamed},
  journal={International Journal of Academic Research in Management (IJARM)},
  volume={11},
  number={1},
  pages={39--51},
  year={2022}
}

@phdthesis{kumar2014knowledge,
  title={Knowledge Discovery Through Spoken Dialog},
  author={Kumar Pappu, Aasish},
  year={2014},
  school={Carnegie Mellon University}
}

@inproceedings{tang2024minicheck,
  title={MiniCheck: Efficient Fact-Checking of LLMs on Grounding Documents},
  author={Tang, Liyan and Laban, Philippe and Durrett, Greg},
  booktitle={Proceedings of the 2024 Conference on Empirical Methods in Natural Language Processing},
  pages={8818--8847},
  year={2024}
}

@inproceedings{de2025supernotes,
  title={Supernotes: Driving consensus in crowd-sourced fact-checking},
  author={De, Soham and Bakker, Michiel A and Baxter, Jay and Saveski, Martin},
  booktitle={Proceedings of the ACM on Web Conference 2025},
  pages={3751--3761},
  year={2025}
}

@inproceedings{deng2023goal,
  title={Goal Awareness for Conversational AI: Proactivity, Non-collaborativity, and Beyond},
  author={Deng, Yang and Lei, Wenqiang and Huang, Minlie and Chua, Tat-Seng},
  booktitle={Proceedings of the 61st Annual Meeting of the Association for Computational Linguistics (Volume 6: Tutorial Abstracts)},
  pages={1--10},
  year={2023}
}

@inproceedings{hong2025augmenting,
  title={Augmenting Compliance-Guaranteed Customer Service Chatbots: Context-Aware Knowledge Expansion with Large Language Models},
  author={Hong, Mengze and Zhang, Chen Jason and Jiang, Di and He, Yuanqin},
  booktitle={Proceedings of the 2025 Conference on Empirical Methods in Natural Language Processing: Industry Track},
  pages={753--765},
  year={2025}
}

@inproceedings{hong2025dial,
  title={Dial-in LLM: Human-aligned LLM-in-the-loop intent clustering for customer service dialogues},
  author={Hong, Mengze and Ng, Wailing and Zhang, Chen Jason and Song, Yuanfeng and Jiang, Di},
  booktitle={Proceedings of the 2025 Conference on Empirical Methods in Natural Language Processing},
  pages={5896--5911},
  year={2025}
}

@inproceedings{song2021emotional,
  title={An emotional comfort framework for improving user satisfaction in E-commerce customer service chatbots},
  author={Song, Shuangyong and Wang, Chao and Chen, Haiqing and Chen, Huan},
  booktitle={Proceedings of the 2021 Conference of the North American Chapter of the Association for Computational Linguistics: Human Language Technologies: Industry Papers},
  pages={130--137},
  year={2021}
}

@inproceedings{yu-etal-2020-conversation,
    title = "When and Who? Conversation Transition Based on Bot-Agent Symbiosis Learning Network",
    author = "Yu, Yipeng  and
      Guan, Ran  and
      Ma, Jie  and
      Jiang, Zhuoxuan  and
      Huang, Jingchang",
    editor = "Scott, Donia  and
      Bel, Nuria  and
      Zong, Chengqing",
    booktitle = "Proceedings of the 28th International Conference on Computational Linguistics",
    month = dec,
    year = "2020",
    address = "Barcelona, Spain (Online)",
    publisher = "International Committee on Computational Linguistics",
    url = "https://aclanthology.org/2020.coling-main.358/",
    doi = "10.18653/v1/2020.coling-main.358",
    pages = "4056--4066",
}

@inproceedings{zheng-etal-2023-dialogqae,
    title = "{D}ialog{QAE}: N-to-N Question Answer Pair Extraction from Customer Service Chatlog",
    author = "Zheng, Xin  and
      Liu, Tianyu  and
      Meng, Haoran  and
      Wang, Xu  and
      Jiang, Yufan  and
      Rao, Mengliang  and
      Lin, Binghuai  and
      Cao, Yunbo  and
      Sui, Zhifang",
    editor = "Bouamor, Houda  and
      Pino, Juan  and
      Bali, Kalika",
    booktitle = "Findings of the Association for Computational Linguistics: EMNLP 2023",
    month = dec,
    year = "2023",
    address = "Singapore",
    publisher = "Association for Computational Linguistics",
    url = "https://aclanthology.org/2023.findings-emnlp.435/",
    doi = "10.18653/v1/2023.findings-emnlp.435",
    pages = "6540--6558",
}

@inproceedings{yang2023improving,
  title={Improving knowledge production efficiency with question answering on conversation},
  author={Yang, Changlin and Liu, Siye and Hu, Sen and Zhang, Wangshu and Xu, Teng and Zheng, Jing},
  booktitle={Proceedings of the 61st Annual Meeting of the Association for Computational Linguistics (Volume 5: Industry Track)},
  pages={225--234},
  year={2023}
}

@inproceedings{cui-etal-2017-superagent,
    title = "{S}uper{A}gent: A Customer Service Chatbot for {E}-commerce Websites",
    author = "Cui, Lei  and
      Huang, Shaohan  and
      Wei, Furu  and
      Tan, Chuanqi  and
      Duan, Chaoqun  and
      Zhou, Ming",
    editor = "Bansal, Mohit  and
      Ji, Heng",
    booktitle = "Proceedings of {ACL} 2017, System Demonstrations",
    month = jul,
    year = "2017",
    address = "Vancouver, Canada",
    publisher = "Association for Computational Linguistics",
    url = "https://aclanthology.org/P17-4017/",
    pages = "97--102"
}

@inproceedings{sun-etal-2022-tracking,
    title = "Tracking Satisfaction States for Customer Satisfaction Prediction in {E}-commerce Service Chatbots",
    author = "Sun, Yang  and
      Wu, Liangqing  and
      Song, Shuangyong  and
      Yu, Xiaoguang  and
      He, Xiaodong  and
      Fu, Guohong",
    editor = "Calzolari, Nicoletta  and
      Huang, Chu-Ren  and
      Kim, Hansaem  and
      Pustejovsky, James  and
      Wanner, Leo  and
      Choi, Key-Sun  and
      Ryu, Pum-Mo  and
      Chen, Hsin-Hsi  and
      Donatelli, Lucia  and
      Ji, Heng  and
      Kurohashi, Sadao  and
      Paggio, Patrizia  and
      Xue, Nianwen  and
      Kim, Seokhwan  and
      Hahm, Younggyun  and
      He, Zhong  and
      Lee, Tony Kyungil  and
      Santus, Enrico  and
      Bond, Francis  and
      Na, Seung-Hoon",
    booktitle = "Proceedings of the 29th International Conference on Computational Linguistics",
    month = oct,
    year = "2022",
    address = "Gyeongju, Republic of Korea",
    publisher = "International Committee on Computational Linguistics",
    url = "https://aclanthology.org/2022.coling-1.51/",
    pages = "616--625",
}

@article{agarwal2021evaluating,
  title={Evaluating empathetic chatbots in customer service settings},
  author={Agarwal, Akshay and Maiya, Shashank and Aggarwal, Sonu},
  journal={arXiv preprint arXiv:2101.01334},
  year={2021}
}

@article{vasquez2021one,
  title={One system to rule them all: A universal intent recognition system for customer service chatbots},
  author={Vasquez-Correa, Juan Camilo and Guerrero-Sierra, Juan Carlos and Pemberty-Tamayo, Jose Luis and Jaramillo, Juan Esteban and Tejada-Castro, Andres Felipe},
  journal={arXiv preprint arXiv:2112.08261},
  year={2021}
}

@article{mashaabi2022natural,
  title={Natural language processing in customer service: A systematic review},
  author={Mashaabi, Malak and Alotaibi, Areej and Qudaih, Hala and Alnashwan, Raghad and Al-Khalifa, Hend},
  journal={arXiv preprint arXiv:2212.09523},
  year={2022}
}

@inproceedings{zhang-etal-2024-clamber,
    title = "{CLAMBER}: A Benchmark of Identifying and Clarifying Ambiguous Information Needs in Large Language Models",
    author = "Zhang, Tong  and
      Qin, Peixin  and
      Deng, Yang  and
      Huang, Chen  and
      Lei, Wenqiang  and
      Liu, Junhong  and
      Jin, Dingnan  and
      Liang, Hongru  and
      Chua, Tat-Seng",
    editor = "Ku, Lun-Wei  and
      Martins, Andre  and
      Srikumar, Vivek",
    booktitle = "Proceedings of the 62nd Annual Meeting of the Association for Computational Linguistics (Volume 1: Long Papers)",
    month = aug,
    year = "2024",
    address = "Bangkok, Thailand",
    publisher = "Association for Computational Linguistics",
    url = "https://aclanthology.org/2024.acl-long.578/",
    doi = "10.18653/v1/2024.acl-long.578",
    pages = "10746--10766",
}

@inproceedings{chen-etal-2024-style,
    title = "{STYLE}: Improving Domain Transferability of Asking Clarification Questions in Large Language Model Powered Conversational Agents",
    author = "Chen, Yue  and
      Huang, Chen  and
      Deng, Yang  and
      Lei, Wenqiang  and
      Jin, Dingnan  and
      Liu, Jia  and
      Chua, Tat-Seng",
    editor = "Ku, Lun-Wei  and
      Martins, Andre  and
      Srikumar, Vivek",
    booktitle = "Findings of the Association for Computational Linguistics: ACL 2024",
    month = aug,
    year = "2024",
    address = "Bangkok, Thailand",
    publisher = "Association for Computational Linguistics",
    url = "https://aclanthology.org/2024.findings-acl.632/",
    doi = "10.18653/v1/2024.findings-acl.632",
    pages = "10633--10649",
}

@inproceedings{10.1145/3269206.3271776,
author = {Zhang, Yongfeng and Chen, Xu and Ai, Qingyao and Yang, Liu and Croft, W. Bruce},
title = {Towards Conversational Search and Recommendation: System Ask, User Respond},
year = {2018},
isbn = {9781450360142},
publisher = {Association for Computing Machinery},
address = {New York, NY, USA},
url = {https://doi.org/10.1145/3269206.3271776},
doi = {10.1145/3269206.3271776},
booktitle = {Proceedings of the 27th ACM International Conference on Information and Knowledge Management},
pages = {177–186},
numpages = {10},
location = {Torino, Italy},
series = {CIKM '18}
}

@inproceedings{10.1145/3485447.3512088,
author = {Zhang, Yiming and Wu, Lingfei and Shen, Qi and Pang, Yitong and Wei, Zhihua and Xu, Fangli and Long, Bo and Pei, Jian},
title = {Multiple Choice Questions based Multi-Interest Policy Learning for Conversational Recommendation},
year = {2022},
isbn = {9781450390965},
publisher = {Association for Computing Machinery},
address = {New York, NY, USA},
url = {https://doi.org/10.1145/3485447.3512088},
doi = {10.1145/3485447.3512088},
booktitle = {Proceedings of the ACM Web Conference 2022},
pages = {2153–2162},
numpages = {10},
location = {Virtual Event, Lyon, France},
series = {WWW '22}
}

@inproceedings{chen2023travel,
  title={Travel: Tag-aware conversational faq retrieval via reinforcement learning},
  author={Chen, Yue and Jin, Dingnan and Huang, Chen and Liu, Jia and Lei, Wenqiang},
  booktitle={Proceedings of the 2023 Conference on Empirical Methods in Natural Language Processing},
  pages={3861--3872},
  year={2023}
}

@article{wu-etal-2023-inscit,
    title = "{I}n{SCI}t: Information-Seeking Conversations with Mixed-Initiative Interactions",
    author = "Wu, Zeqiu  and
      Parish, Ryu  and
      Cheng, Hao  and
      Min, Sewon  and
      Ammanabrolu, Prithviraj  and
      Ostendorf, Mari  and
      Hajishirzi, Hannaneh",
    journal = "Transactions of the Association for Computational Linguistics",
    volume = "11",
    year = "2023",
    address = "Cambridge, MA",
    publisher = "MIT Press",
    url = "https://aclanthology.org/2023.tacl-1.27/",
    doi = "10.1162/tacl_a_00559",
    pages = "453--468"
}

@inproceedings{min-etal-2019-multi,
    title = "Multi-hop Reading Comprehension through Question Decomposition and Rescoring",
    author = "Min, Sewon  and
      Zhong, Victor  and
      Zettlemoyer, Luke  and
      Hajishirzi, Hannaneh",
    editor = "Korhonen, Anna  and
      Traum, David  and
      M{\`a}rquez, Llu{\'i}s",
    booktitle = "Proceedings of the 57th Annual Meeting of the Association for Computational Linguistics",
    month = jul,
    year = "2019",
    address = "Florence, Italy",
    publisher = "Association for Computational Linguistics",
    url = "https://aclanthology.org/P19-1613/",
    doi = "10.18653/v1/P19-1613",
    pages = "6097--6109",
}

@incollection{oudejans2018using,
  title={Using interactive features to motivate and probe responses to open-ended questions},
  author={Oudejans, Marije},
  booktitle={Social and behavioral research and the internet},
  pages={215--244},
  year={2018},
  publisher={Routledge}
}

@article{nicolescu2022human,
  title={Human-computer interaction in customer service: the experience with AI chatbots—a systematic literature review},
  author={Nicolescu, Luminița and Tudorache, Monica Teodora},
  journal={Electronics},
  volume={11},
  number={10},
  pages={1579},
  year={2022},
  publisher={Mdpi}
}

@article{adam2021ai,
  title={AI-based chatbots in customer service and their effects on user compliance},
  author={Adam, Martin and Wessel, Michael and Benlian, Alexander},
  journal={Electronic markets},
  volume={31},
  number={2},
  pages={427--445},
  year={2021},
  publisher={Springer}
}

@inproceedings{liu2025learning,
  title={Learning to refuse: Towards mitigating privacy risks in llms},
  author={Liu, Zhenhua and Zhu, Tong and Tan, Chuanyuan and Chen, Wenliang},
  booktitle={Proceedings of the 31st International Conference on Computational Linguistics},
  pages={1683--1698},
  year={2025}
}

@book{sarstedt2014concise,
  title={A concise guide to market research},
  author={Sarstedt, Marko and Mooi, Erik and others},
  volume={324},
  year={2014},
  publisher={Springer}
}

@book{bradburn2004asking,
  title={Asking questions: the definitive guide to questionnaire design--for market research, political polls, and social and health questionnaires},
  author={Bradburn, Norman M and Sudman, Seymour and Wansink, Brian},
  year={2004},
  publisher={John Wiley \& Sons}
}

@article{lenzner2010cognitive,
  title={Cognitive burden of survey questions and response times: A psycholinguistic experiment},
  author={Lenzner, Timo and Kaczmirek, Lars and Lenzner, Alwine},
  journal={Applied cognitive psychology},
  volume={24},
  number={7},
  pages={1003--1020},
  year={2010},
  publisher={Wiley Online Library}
}

@inproceedings{zhang-etal-2023-towards-effective,
    title = "Towards Effective Automatic Debt Collection with Persona Awareness",
    author = "Zhang, Tong  and
      Liu, Junhong  and
      Huang, Chen  and
      Liu, Jia  and
      Liang, Hongru  and
      Wen, Zujie  and
      Lei, Wenqiang",
    editor = "Wang, Mingxuan  and
      Zitouni, Imed",
    booktitle = "Proceedings of the 2023 Conference on Empirical Methods in Natural Language Processing: Industry Track",
    month = dec,
    year = "2023",
    address = "Singapore",
    publisher = "Association for Computational Linguistics",
    url = "https://aclanthology.org/2023.emnlp-industry.4/",
    doi = "10.18653/v1/2023.emnlp-industry.4",
    pages = "32--45",
}

@Article{info15080432,
AUTHOR = {Ridzuan, Nurhadhinah Nadiah and Masri, Masairol and Anshari, Muhammad and Fitriyani, Norma Latif and Syafrudin, Muhammad},
TITLE = {AI in the Financial Sector: The Line between Innovation, Regulation and Ethical Responsibility},
JOURNAL = {Information},
VOLUME = {15},
YEAR = {2024},
NUMBER = {8},
ARTICLE-NUMBER = {432},
URL = {https://www.mdpi.com/2078-2489/15/8/432},
ISSN = {2078-2489},
DOI = {10.3390/info15080432}
}

@article{behr2012asking,
  title={Asking probing questions in web surveys: which factors have an impact on the quality of responses?},
  author={Behr, Doroth{\'e}e and Kaczmirek, Lars and Bandilla, Wolfgang and Braun, Michael},
  journal={Social Science Computer Review},
  volume={30},
  number={4},
  pages={487--498},
  year={2012},
  publisher={Sage Publications Sage CA: Los Angeles, CA}
}

@inproceedings{10.1145/3706598.3714128,
author = {Jacobsen, Rune M\o{}berg and Cox, Samuel Rhys and Griggio, Carla F. and van Berkel, Niels},
title = {Chatbots for Data Collection in Surveys: A Comparison of Four Theory-Based Interview Probes},
year = {2025},
isbn = {9798400713941},
publisher = {Association for Computing Machinery},
address = {New York, NY, USA},
url = {https://doi.org/10.1145/3706598.3714128},
doi = {10.1145/3706598.3714128},
booktitle = {Proceedings of the 2025 CHI Conference on Human Factors in Computing Systems},
articleno = {228},
numpages = {21},
location = {
},
series = {CHI '25}
}

@inproceedings{deng2023survey,
  title={A survey on proactive dialogue systems: problems, methods, and prospects},
  author={Deng, Yang and Lei, Wenqiang and Lam, Wai and Chua, Tat-Seng},
  booktitle={Proceedings of the Thirty-Second International Joint Conference on Artificial Intelligence},
  pages={6583--6591},
  year={2023}
}

@inproceedings{singh-rios-2022-linguistic,
    title = "Linguistic Elements of Engaging Customer Service Discourse on Social Media",
    author = "Singh, Sonam  and
      Rios, Anthony",
    editor = "Bamman, David  and
      Hovy, Dirk  and
      Jurgens, David  and
      Keith, Katherine  and
      O'Connor, Brendan  and
      Volkova, Svitlana",
    booktitle = "Proceedings of the Fifth Workshop on Natural Language Processing and Computational Social Science (NLP+CSS)",
    month = nov,
    year = "2022",
    address = "Abu Dhabi, UAE",
    publisher = "Association for Computational Linguistics",
    url = "https://aclanthology.org/2022.nlpcss-1.12/",
    doi = "10.18653/v1/2022.nlpcss-1.12",
    pages = "105--117",
}

@inproceedings{hsu-etal-2024-evaluating,
    title = "Evaluating {C}hat{N}et{Z}ero, an {LLM}-Chatbot to Demystify Climate Pledges",
    author = "Hsu, Angel  and
      Laney, Mason  and
      Zhang, Ji  and
      Manya, Diego  and
      Farczadi, Linda",
    editor = "Stammbach, Dominik  and
      Ni, Jingwei  and
      Schimanski, Tobias  and
      Dutia, Kalyan  and
      Singh, Alok  and
      Bingler, Julia  and
      Christiaen, Christophe  and
      Kushwaha, Neetu  and
      Muccione, Veruska  and
      A. Vaghefi, Saeid  and
      Leippold, Markus",
    booktitle = "Proceedings of the 1st Workshop on Natural Language Processing Meets Climate Change (ClimateNLP 2024)",
    month = aug,
    year = "2024",
    address = "Bangkok, Thailand",
    publisher = "Association for Computational Linguistics",
    url = "https://aclanthology.org/2024.climatenlp-1.6/",
    doi = "10.18653/v1/2024.climatenlp-1.6",
    pages = "82--92",
}

@inproceedings{taranukhin-etal-2024-empowering,
    title = "Empowering Air Travelers: A Chatbot for {C}anadian Air Passenger Rights",
    author = "Taranukhin, Maksym  and
      Ravi, Sahithya  and
      Lukacs, Gabor  and
      Milios, Evangelos  and
      Shwartz, Vered",
    editor = "Aletras, Nikolaos  and
      Chalkidis, Ilias  and
      Barrett, Leslie  and
      Goanț{\u{a}}, C{\u{a}}t{\u{a}}lina  and
      Preoțiuc-Pietro, Daniel  and
      Spanakis, Gerasimos",
    booktitle = "Proceedings of the Natural Legal Language Processing Workshop 2024",
    month = nov,
    year = "2024",
    address = "Miami, FL, USA",
    publisher = "Association for Computational Linguistics",
    url = "https://aclanthology.org/2024.nllp-1.27/",
    doi = "10.18653/v1/2024.nllp-1.27",
    pages = "326--335",
}

@inproceedings{li-etal-2025-ai,
    title = "{AI} Chatbots as Professional Service Agents: Developing a Professional Identity",
    author = "Li, Wenwen  and
      Shi, Kangwei  and
      Chai, Yidong",
    editor = "Christodoulopoulos, Christos  and
      Chakraborty, Tanmoy  and
      Rose, Carolyn  and
      Peng, Violet",
    booktitle = "Proceedings of the 2025 Conference on Empirical Methods in Natural Language Processing",
    month = nov,
    year = "2025",
    address = "Suzhou, China",
    publisher = "Association for Computational Linguistics",
    url = "https://aclanthology.org/2025.emnlp-main.652/",
    doi = "10.18653/v1/2025.emnlp-main.652",
    pages = "12923--12936",
    ISBN = "979-8-89176-332-6",
}

@inproceedings{guo-etal-2025-candy,
    title = "{CANDY}: Benchmarking {LLM}s' Limitations and Assistive Potential in {C}hinese Misinformation Fact-Checking",
    author = "Guo, Ruiling  and
      Yang, Xinwei  and
      Huang, Chen  and
      Zhang, Tong  and
      Hu, Yong",
    editor = "Christodoulopoulos, Christos  and
      Chakraborty, Tanmoy  and
      Rose, Carolyn  and
      Peng, Violet",
    booktitle = "Findings of the Association for Computational Linguistics: EMNLP 2025",
    month = nov,
    year = "2025",
    address = "Suzhou, China",
    publisher = "Association for Computational Linguistics",
    url = "https://aclanthology.org/2025.findings-emnlp.307/",
    doi = "10.18653/v1/2025.findings-emnlp.307",
    pages = "5724--5758",
    ISBN = "979-8-89176-335-7",
}

@inproceedings{
cheng2024adapting,
title={Adapting Large Language Models via Reading Comprehension},
author={Daixuan Cheng and Shaohan Huang and Furu Wei},
booktitle={The Twelfth International Conference on Learning Representations},
year={2024},
url={https://openreview.net/forum?id=y886UXPEZ0}
}

@inproceedings{chen-etal-2021-finqa,
    title = "{F}in{QA}: A Dataset of Numerical Reasoning over Financial Data",
    author = "Chen, Zhiyu  and
      Chen, Wenhu  and
      Smiley, Charese  and
      Shah, Sameena  and
      Borova, Iana  and
      Langdon, Dylan  and
      Moussa, Reema  and
      Beane, Matt  and
      Huang, Ting-Hao  and
      Routledge, Bryan  and
      Wang, William Yang",
    editor = "Moens, Marie-Francine  and
      Huang, Xuanjing  and
      Specia, Lucia  and
      Yih, Scott Wen-tau",
    booktitle = "Proceedings of the 2021 Conference on Empirical Methods in Natural Language Processing",
    month = nov,
    year = "2021",
    address = "Online and Punta Cana, Dominican Republic",
    publisher = "Association for Computational Linguistics",
    url = "https://aclanthology.org/2021.emnlp-main.300/",
    doi = "10.18653/v1/2021.emnlp-main.300",
    pages = "3697--3711",
}

@misc{fu2023improving,
      title={Improving Language Model Negotiation with Self-Play and In-Context Learning from AI Feedback}, 
      author={Yao Fu and Hao Peng and Tushar Khot and Mirella Lapata},
      year={2023},
      eprint={2305.10142},
      archivePrefix={arXiv},
      primaryClass={cs.CL}
}

@misc{yang2025qwen3technicalreport,
      title={Qwen3 Technical Report}, 
      author={An Yang and Anfeng Li and Baosong Yang and Beichen Zhang and Binyuan Hui and Bo Zheng and Bowen Yu and Chang Gao and Chengen Huang and Chenxu Lv and Chujie Zheng and Dayiheng Liu and Fan Zhou and Fei Huang and Feng Hu and Hao Ge and Haoran Wei and Huan Lin and Jialong Tang and Jian Yang and Jianhong Tu and Jianwei Zhang and Jianxin Yang and Jiaxi Yang and Jing Zhou and Jingren Zhou and Junyang Lin and Kai Dang and Keqin Bao and Kexin Yang and Le Yu and Lianghao Deng and Mei Li and Mingfeng Xue and Mingze Li and Pei Zhang and Peng Wang and Qin Zhu and Rui Men and Ruize Gao and Shixuan Liu and Shuang Luo and Tianhao Li and Tianyi Tang and Wenbiao Yin and Xingzhang Ren and Xinyu Wang and Xinyu Zhang and Xuancheng Ren and Yang Fan and Yang Su and Yichang Zhang and Yinger Zhang and Yu Wan and Yuqiong Liu and Zekun Wang and Zeyu Cui and Zhenru Zhang and Zhipeng Zhou and Zihan Qiu},
      year={2025},
      eprint={2505.09388},
      archivePrefix={arXiv},
      primaryClass={cs.CL},
      url={https://arxiv.org/abs/2505.09388}, 
}

@article{deng2025proactive,
  title={Proactive conversational ai: A comprehensive survey of advancements and opportunities},
  author={Deng, Yang and Liao, Lizi and Lei, Wenqiang and Yang, Grace Hui and Lam, Wai and Chua, Tat-Seng},
  journal={ACM Transactions on Information Systems},
  volume={43},
  number={3},
  pages={1--45},
  year={2025},
  publisher={ACM New York, NY}
}

\appendix

\section{Implementation Details}
\label{impl}
All training experiments are conducted on a server equipped with a single NVIDIA A6000 GPU.

\subsection{Implementation Details of Baselines}
All baseline methods are re-implemented by us according to the original papers and supplementary materials, with careful effort to match the reported settings as closely as possible.
\begin{itemize}[leftmargin=*]
    \item \textbf{Vanilla}: This baseline functions as a conventional customer service chatbot. It is not equipped with any specific prompts or strategies for probing; instead, it is restricted to interacting with the user in a natural, purely reactive manner.
    
    \item \textbf{Proactive}: Following \cite{deng2023prompting}, we prompt the LLM to analyze the dialogue history and target information, plan the next-step strategy, and generate a response conditioned on the planned strategy and target information. The prompt design is shown in Table \ref{tab:proactive_prompt}.


    \item \textbf{ICL-AIF}: Following \cite{fu2023improving}, at each dialogue turn, we prompt the LLM to first generate three action-level suggestions for the next response based on the accumulated dialogue history and target information. Conditioned on these suggestions, the LLM then generates the response for the current turn. The prompt design is shown in Table \ref{tab:iclaif_prompt}.
\end{itemize}
\subsection{Implementation Details of \ours}
\begin{figure*}
    \centering
    \includegraphics[width=0.98\linewidth]{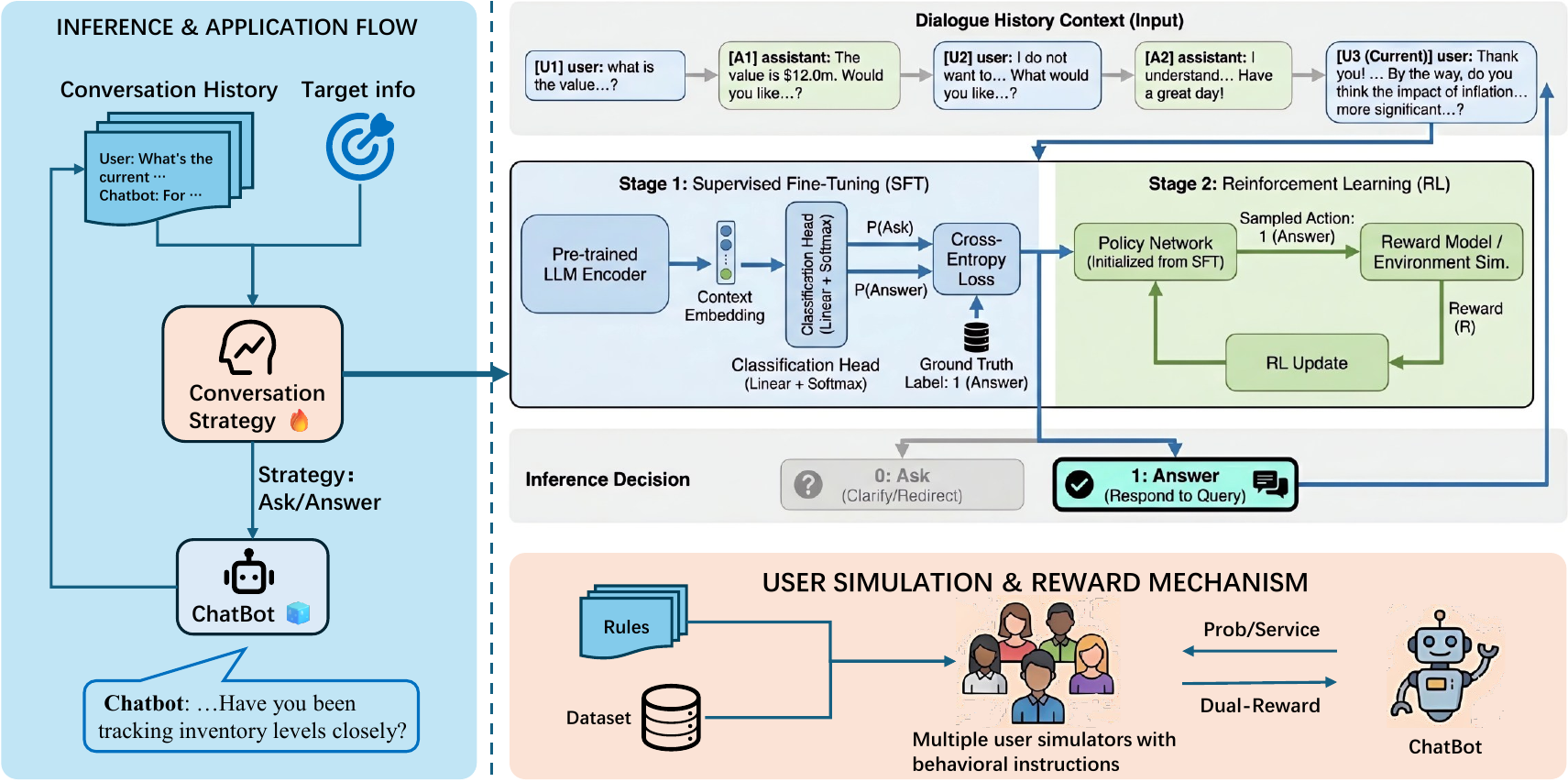}
    \caption{Overview of \ours. It incorporates a conversation strategy module (CS) that explicitly guides the timing of information probe $d_t$. Here, the CS module is trained via two-stage curriculum.}
    \label{fig:method2}
\end{figure*}

Figure \ref{fig:method2} details the data flow of \ours.

\subsubsection{Supervised Fine-Tuning}
\noindent\textbf{Rule-based Data Generation}. We initialize the CS module using synthetic, labeled dialogues generated by LLMs to train the probing strategy, with example dialogues shown in Table \ref{tab:sft_example}. To achieve this, we generate a set of governing rules that define appropriate and inappropriate probing scenarios. Refer to Table \ref{tab:rules} for these governing rules. Finally, we utilize GPT-4o Mini to generate $8$ governing rules and construct a final dataset $\mathcal{D}_{\text{sft}}$ containing $600$ dialogue sessions. 

\begin{table*}[h]
\centering
\begin{tabular}{|l|p{\linewidth-3cm}|}
\hline
\textbf{Text} & user: What is the net change from 2013 to 2015? \newline
assistant: The net change from 2013 to 2015 in the capital structure is reflected in changes in common stockholders' equity, long-term debt, and short-term debt. Would you like to know the specific net change in any of these components? \newline
user: I'd like to see the breakdown of the net change in common stockholders' equity. Also, could you provide details on the company's revenue growth during those years? \\
\hline
\textbf{Label} & Ask \\
\hline
\end{tabular}
\caption{Labeled Dialogue Examples for SFT Data Construction}
\label{tab:sft_example}
\end{table*}

\begin{table*}[h]
\centering
\begin{tabular}{|l|p{\linewidth-3cm}|}
\hline
\textbf{ask} & User provides market information but only partially (e.g., price, trend, supply, policy). \newline
User language indicates ongoing attention or experience with the market. \newline
User shows willingness to shift or expand the topic. \newline
The user’s previous question has been answered. \\
\hline
\textbf{answer} & User explicitly refuses or avoids answering. \newline
User has already provided sufficient, multi-dimensional market information. \newline
Further questions would elicit opinions or interpretations rather than facts. \newline
Conversation focus shifts to the user asking questions. \\
\hline
\end{tabular}
\caption{Governing Rules}
\label{tab:rules}
\end{table*}

\noindent\textbf{SFT}. The probing classifier is implemented using Hugging Face Transformers' \texttt{AutoModelForSequenceClassification} and trained via supervised fine-tuning (SFT). The training objective minimizes the binary cross-entropy loss between the predicted probing decision \(\hat{d}_t\) and the ground-truth label \(d_t\):
\begin{equation}
\mathcal{L}_{\text{CE}} = - \frac{1}{m} \sum_{i=1}^{m} \big[d_i \log \hat{d}_i + (1 - d_i) \log (1 - \hat{d}_i)\big].
\end{equation}
We train the model with a batch size of 5 and a learning rate of \(2\times10^{-5}\), using the AdamW optimizer with no weight decay. The checkpoint achieving the best performance on the validation set is selected for evaluation.

\subsubsection{Online RL Training}
After SFT, we further optimize the CS module using the REINFORCE algorithm. In our setup, training involves 500 episodes, with a learning rate of \(1\times10^{-5}\), a discount factor of 0.99, and a maximum of 8 dialogue turns per episode. 

\noindent\textbf{Environment/User Simulator}. To simulate user resistance, we construct a set of behavioral instructions \(\mathcal{K}\) describing conditions under which the user may refuse to answer or reduce satisfaction. These instructions include:

\begin{itemize}[leftmargin=*]
    \item An expert repeatedly asks you questions
    \item The question is unrelated to the current topic
    \item It interrupts your expression
    \item The question is impolite or overly complicated
    \item The timing is inappropriate or conflicts with the context
    \item The question is too complex to answer
    \item The question involves private or sensitive information
    \item You don't trust the other party yet and don't want to answer
\end{itemize}

\noindent We consider all combinations of 1 to 7 instructions from this set, resulting in 254 unique combinations. These are split evenly into a training set and a test set, each containing 127 combinations, ensuring no overlap between training and testing scenarios. During each training iteration, a simulator is randomly selected from \(\mathcal{S}\) and a random instruction subset from \(\mathcal{K}\) is applied to guide its responses, exposing \ours~to diverse resistance patterns and promoting a robust probing strategy.

\noindent\textbf{Reward Design}. For reward design of each dataset, we set +2.0 for Success probe, -1.0 for Rejected probe, -0.2 for Invalid Probe, +1.5 for Smart Stop, +0.3 for Mitigation, and -2.0 for Passive Penalty. For all other cases, the reward is 0.0. 

\subsubsection{Strategy Prompting}
Here, we describe the prompts used in our experiments. Different strategies correspond to different prompt formulations, all of which are summarized as our \ours~prompting in Table \ref{tab:our_prompt}.

\section{Evaluation Details}
\subsection{Datasets}
We evaluate our method across two datasets, including FinQA \cite{chen-etal-2021-finqa}, which features complex multi-hop questions, and ConvQA \cite{cheng2024adapting}, which is an extension of FinQA designed for conversational contexts. It breaks down complex reasoning chains into sequences of simplified, decomposed questions, mimicking a natural dialogue flow.

In our experiments, we preprocess each dataset using the same pipeline. For both FinQA and ConvQA, we first extract the question and its associated context from each example. All questions are then embedded using \texttt{qwen3-embedding-0.6b}, and pairwise cosine similarities are computed. To ensure coverage of question pairs with varying degrees of semantic relatedness and to consider our financial cost, we sample question pairs from different cosine similarity ranges. For FinQA, we randomly sample 400 question pairs with similarity scores $\geq 0.80$, 400 pairs with scores in $[0.60, 0.80)$, and 400 pairs with scores $< 0.60$. For ConvQA, we randomly sample 400 question pairs with similarity scores $\geq 0.80$, 400 pairs with scores in $[0.65, 0.80)$, and 400 pairs with scores $< 0.65$. In total, this results in 1,200 paired instances for FinQA and 1,200 paired instances for ConvQA, which constitute the final datasets used in our experiments.

Notably, these datasets serve solely to initialize the user queries and the chatbot's target information $\mathcal{I}$. Specifically, for each dataset, we partition the questions into two disjoint, equal-sized subsets. We then randomly sample one question from each subset to form a pair ($q_1$, $q_2$), where $q_1$ is designated as the user's inquiry, while $q_2$ serves as the target information the chatbot must probe. To ensure a grounded interaction, we provide the corresponding context associated with $q_1$ to the chatbot, and the context for $q_2$ to the user.

\subsection{Metrics}
We employ the following metrics, which provide a comprehensive assessment of both the information acquisition and service quality.

\begin{itemize}[leftmargin=*]
    \item \textbf{Task Success Rate} (TSR) measures the efficacy of information gathering. It is defined as the percentage of dialogue sessions in which the chatbot successfully acquires the target information $\mathcal{I}$ through probing. Formally, 
    \begin{equation}
        \text{TSR} = \frac{N_{\text{success}}}{N_{\text{total}}},
    \end{equation}
    where $N_{\text{success}}$ is the number of sessions where target information is acquired. Notably, we employ both LLM-as-a-judge (or human, depended on who is the evaluator) to determine whether the target information is successfully elicited.
    
    \item \textbf{Average Turns} (AvgT) evaluate interaction efficiency. We report the average number of dialogue turns per session. Lower values indicate more efficient probing strategies that minimize user effort. Formally,  
    \begin{equation}
        \text{AvgT} = \frac{1}{N_{\text{total}}} \sum_{i} T_i,
    \end{equation}
    where $T_i$ is the turn count of session $i$. Lower values indicate reduced user effort. 

    \item \textbf{Rejected Probing Rate} (RPR) quantifies user friction. It is calculated as the average proportion of probing attempts within a session that are explicitly rejected or ignored by the user. A lower RPR indicates a less intrusive probing strategy. Formally,
    \begin{equation}
        \text{RPR} = \frac{N_{\text{rejected}}}{N_{\text{probes}}},
    \end{equation}
    where $N_{\text{rejected}}$ is the count of rejected/ignored probes and $N_{\text{probes}}$ is the total probes attempted. Notably, we employ both LLM-as-a-judge (or human, depended on who is the evaluator) to identify probe occurrences and determine whether the user accepted or rejected the inquiry.
    
    \item \textbf{Probing Coherence} (PC) assess the naturalness of the interaction by measuring the contextual consistency of the probes. Scored on a scale of [1-5] by LLM judges (or humans), this metric evaluates whether the inserted probes flow logically from the ongoing dialogue context without abrupt transitions.
    
    \item \textbf{Query Response Rate} (QRR) ensure the chatbot maintains its primary service function, we measure the rate at which user inquiries are addressed. This is defined as the percentage of user queries that receive a responsive reply from the chatbot, regardless of the factual correctness of the answer. Formally, 
    \begin{equation}
        \text{QRR} = \frac{N_{\text{answered}}}{N_{\text{queries}}},
    \end{equation}
    where $N_{\text{answered}}$ is the count of user inquiries addressed by the chatbot. Notably, we employ both LLM-as-a-judge (or human, depended on who is the evaluator) to identify user query occurrences and determine whether the chatbot answers the user quiry.
    
    \item \textbf{Overall User Satisfaction} (OUS) A comprehensive metric reflecting the user's subjective experience. It is measured on a score of 1-5 at the end of each session. To achieve this, we simply denote $OUS = (PC + 5*(1-RPR))/2$. Taking human evaluation for example, OUS is calculated by human-assessed PC score and human-assessed RPR, where rejected probes is counted by humans. Notably, this heuristic provides a composite score that balances conversational quality with user receptiveness. Basically, this formula serves as a quantitative proxy for subjective user satisfaction. It is based on the intuition that a positive user experience is a function of both high conversational quality (PC) and a high degree of user acceptance of the chatbot's proactive probes
\end{itemize}

\subsection{LLM Evaluator}

We use a Gemini-3-flash-preview as an automatic evaluator to assess dialogue behaviors in our experiments. 

The evaluator first performs two types of assessments: (1) response-level assessments, which determine the appropriateness of individual turns and are used to calculate the Rejection Rate (RPR); and (2) dialogue-level coherence scoring, which evaluates the overall conversational flow and serves as the basis for the Probing Coherence (PC) metric. For response-level evaluation, the model conducts two judgments.
The first is an \emph{answer existence judgment} (QR), which determines whether the assistant’s reply answers the user’s question from the previous turn.
The second is a \emph{target satisfaction and refusal judgment} (TS), which determines whether the user’s reply addresses the assistant’s intended target question or explicitly refuses to answer. The evaluation prompts used for QR and TS are provided in Table~\ref{tab:qr_ts_prompt}.

In addition, we assess the coherence of the assistant’s probing behavior using a \emph{Probing Coherence} (PC) score.
PC measures how naturally the assistant’s probes or questions connect to the dialogue history, focusing on transitional phrasing and logical consistency.
It is rated on a five-point ordinal scale, where higher scores indicate smoother and more coherent probing. The prompt used for PC scoring is reported in Table~\ref{tab:pc_prompt}.

We evaluated the consistency between the automatic LLM scoring and human judgments for Probing Coherence (PC). We report both Cohen’s kappa and Krippendorff’s alpha to measure agreement between the LLM evaluator and human annotators. The overall Cohen’s kappa is 0.7282, and the overall Krippendorff’s alpha is 0.6912, indicating substantial agreement and demonstrating that the LLM evaluator can reliably approximate human assessments of probing coherence.

\subsection{Human Evaluators}
To complement the automatic evaluation, a human study was conducted to assess the probing coherence of different methods. Specifically, 10 user simulator dialogues were randomly sampled, and all compared methods were evaluated on the same set of dialogues to ensure fairness.

The human evaluation focused on \textit{Probing Coherence}. Six human evaluators jointly reviewed each dialogue and discussed the probing behavior of the assistant before assigning a final score, following existing setup \cite{deng2023prompting,zhang-etal-2024-clamber,chen-etal-2024-style}. According to \citet{zhang-etal-2023-towards-effective}, the discussion-based protocol helps reduce individual subjectivity and to encourage consensus-based judgments. Due to the resource-intensive nature of human assessment, our use of six evaluators aligns with established practices in the field \cite{}. This sample size provides a sufficiently rigorous basis to validate the effectiveness of our proposed method.

The scoring criteria strictly followed the same guidelines used in the large language model–based evaluation. In particular, evaluators assigned scores according to the probing coherence rubric described in Table~\ref{tab:pc_prompt}, considering factors such as semantic relevance between probes and preceding context, smoothness of topic transitions, and the naturalness of information-seeking behavior.

\subsection{Human Participants}
Beyond interactions with user simulators, we conduct a user study involving 6 human participants to interact with the chatbots, following existing setup \cite{deng2023prompting,zhang-etal-2024-clamber,chen-etal-2024-style}. Participants are instructed to role-play as customers (provided with specific queries) and are given full autonomy to decide whether to answer the chatbot's probes based on their own experience. Given the high cost of human studies, we compare \ours~exclusively against the proactive baselines. We employ a controlled, within-subjects design: each participant engage with every method in five separate interactions. To ensure a fair comparison, both the probing tasks for the chatbot and the initial service queries from the user are standardized across all sessions and methods. The order of methods is randomized for each participant to mitigate order effects. \textbf{Unlike the simulators, human participants are instructed to decide if to answer probes based entirely on their own subjective judgment and satisfaction levels.} This configuration ensures more realistic interactions by avoiding the formulaic and may evasive responses—such as 'I do not want to answer right now', encountered in LLM-based user simulators sometimes.

Due to the resource-intensive nature of human participants, our use of six participants also aligns with established practices in the field \cite{zhang-etal-2024-clamber, chen-etal-2024-style}. This sample size provides a sufficiently rigorous basis to validate the effectiveness of our proposed method.

\section{Additional Experiments on Larger LLM Backbone}
We conduct additional experiments using a state-of-the-art large-scale backbone: GPT-5.2. As shown in the Table \ref{tab:large} (evaluated via both Gemini-as-a-judge and computational metrics, same as our paper), the performance trends are consistent with our original findings. This confirms that our proposed method is robust and effective across model scales.

\begin{table}[]
\centering
\resizebox{0.49\textwidth}{!}{%
\begin{tabular}{l|l|l|l|l|l|l}
\toprule
\textbf{Methods} & \textbf{TSR$\uparrow$} & \textbf{AvgT$\downarrow$} & \textbf{RPR$\downarrow$} & \textbf{QRR$\uparrow$} & \textbf{PC} & \textbf{OUS} \\ \midrule
Proactive        & 86.00\%       & 2.79           & 34.33\%       & 92.46\%       & 4.80        & 4.04         \\ \hline
ICL-AIF          & 58.25\%       & 4.60           & 70.16\%       & 99.76\%       & 4.62        & 3.06         \\ \hline
PROCHATIP        & 92.08\%       & 1.57           & 17.50\%       & 76.38\%       & 4.67        & 4.40         \\ \bottomrule
\end{tabular}%
}
\caption{Performance on larger LLM backbone}
\label{tab:large}
\end{table}


\section{Case Studies}
\label{cases}
This section presents qualitative case studies to further analyze the probing behaviors of different methods. Two types of cases are considered: \textit{good cases}, where \ours~demonstrates clear advantages over baseline methods, and \textit{bad cases}, where all methods, including \ours, fail to effectively achieve the probing objective.

In this good case (Table~\ref{tab:good_case_part1}--\ref{tab:good_case_part2}), \ours~neither responds mechanically to user queries nor persistently presses for answers. Instead, it gradually guides the dialogue toward the objective by combining informative responses with strategically timed questions. When the user shows hesitation or temporary refusal, the model sustains the conversation with explanations, preventing disengagement and maintaining continuity. Finally, the probing question is introduced only when the topic naturally aligns with the assistant’s objective, making it coherent and well-motivated, thus increasing the likelihood of eliciting the desired information.

In this bad case (Table~\ref{tab:bad_case_part1}--\ref{tab:bad_case_part2}), the user is highly reluctant to answer and the target information is weakly related to the current topic. Under these conditions, all methods fail to elicit the desired information effectively. Even the most effective strategy can only choose not to press further in order to avoid user rejection. The primary limitation lies in the strategy set, which consists solely of “Ask” and “Answer”; it lacks mechanisms to build trust, smoothly redirect the conversation, or otherwise reduce user resistance, making effective probing infeasible in such challenging scenarios.

\section{Error Analysis}
We focuse on the 100 samples with the lowest OUS (Overall User Satisfaction) scores. Inspired by \citet{zhang-etal-2023-towards-effective}, we adopted an inductive analysis approach. We utilized GPT-5 to diagnose the specific cause of low scores for each sample (identifying up to two reasons per sample). We then performed semantic clustering on these outputs to categorize the failure modes. Finally, a PhD student verified and refined the clustering results.

We identified 4 primary failure categories. The distribution is as follows:
\begin{itemize}[leftmargin=*]
    \item \underline{Identification Failure} (44\%): The assistant failed to accurately capture key information, temporal data, or the user's core intent.
    \item \underline{Goal Inflexibility} (25\%): The assistant fixated on pre-set probing targets while neglecting the user's actual queries or shifts in interest.
    \item \underline{Repetitive Questioning} (17\%): The assistant ignored explicit user refusals and mechanically repeated queries instead of adjusting its strategy.
    \item \underline{Inadequate Steering} (14\%): The assistant struggled to adapt to conversational shifts or effectively steer the user toward the target probing questions.

\end{itemize}

\begin{table*}[ht]
\centering
\begin{tabular}{@{}p{0.95\textwidth}@{}}
\toprule
\textbf{A Good Case of Probing Dialogue (Part 1)} \\
\midrule

\textbf{Behavioral Instructions:} \\
\quad -- An expert repeatedly asks you questions; \\
\quad -- It interrupts your expression. \\[0.5em]

\textbf{User Query:} \\
\quad What was the percentage change in net cash provided by operating activities between 2011 and 2012? \\[0.5em]

\textbf{Chatbot’s Target Information:} \\
\quad In millions for 2013, 2012, and 2011, what was total commercial mortgages? \\[1em]

\textbf{Turn 1 (Strategy: Ask)} \\
\textit{Chatbot:} The percentage change in net cash provided by operating activities between 2011 and 2012 was a decrease of approximately 10.5\%, as it dropped from \$20{,}963 million in 2011 to \$18{,}884 million in 2012. This change highlights the impact of lower net income and working capital adjustments on cash flow, which prompts us to explore how these factors might relate to other financial activities, such as total commercial mortgages. Would you like to know more about the breakdown of cash flows for investing and financing activities in 2013, 2012, and 2011? \\
\textit{User:} Yes, I would like to know more about the breakdown of cash flows for investing and financing activities in those years. \\[0.75em]

\textbf{Turn 2 (Strategy: Ask)} \\
\textit{Chatbot:} The net cash provided by operating activities decreased by approximately 10.5\% from 2011 to 2012, which was influenced by lower net income and changes in working capital. To better understand the overall cash flow dynamics, it would be helpful to examine the total commercial mortgages for 2013, 2012, and 2011, as they can provide insight into how capital allocation and debt management impacted the company's financial position. Would you like me to look up the total commercial mortgages for those years? \\
\textit{User:} I do not want to answer right now. How did the firm's involvement in securitization activities impact its financial performance in 2013? \\

\bottomrule
\end{tabular}
\caption{A Good Case of Probing Dialogue (Part 1)}
\label{tab:good_case_part1}
\end{table*}

\begin{table*}[ht]
\centering
\begin{tabular}{@{}p{0.95\textwidth}@{}}
\toprule
\textbf{A Good Case of Probing Dialogue (Part 2)} \\
\midrule

\textbf{Turn 3 (Strategy: Answer)} \\
\textit{Chatbot:} The firm's involvement in securitization activities in 2013 is not directly mentioned in the provided cash flow data, but the overall net cash provided by operating activities decreased by approximately 10.5\% compared to 2012, while net cash used for investing activities increased significantly. These trends suggest that the firm may have been allocating more capital to long-term investments or acquisitions, which could be related to securitization activities, though further details would be needed to confirm this impact. \\
\textit{User:} What were the main factors contributing to the increase in net cash used for investing activities in 2013? \\[0.75em]

\textbf{Turn 4 (Strategy: Answer)} \\
\textit{Chatbot:} The main factors contributing to the increase in net cash used for investing activities in 2013 were higher purchases of available-for-sale investments and a decrease in maturities and sales of trading assets, partially offset by an increase in maturities and sales of available-for-sale investments and a decrease in purchases of licensed technology and patents. \\
\textit{User:} How did these investment activities align with the firm's strategic goals for 2013? \\[0.75em]

\textbf{Turn 5 (Strategy: Ask)} \\
\textit{Chatbot:} The increase in net cash used for investing activities in 2013 reflects a shift toward higher capital expenditures and investment purchases, which may indicate the firm's focus on expanding its asset base or diversifying its investment portfolio. To better understand how these activities align with the firm's strategic goals, it would be useful to examine the total commercial mortgages for 2013, 2012, and 2011, as they can provide insight into the firm's approach to managing its real estate and financing assets. What were the total commercial mortgages in millions for those years? \\
\textit{User:} The total commercial mortgages were \$6{,}086 million in 2013, \$300 million in 2012, and \$2{,}014 million in 2011. How do you think these figures might relate to the firm's strategic goals? \\

\bottomrule
\end{tabular}
\caption{A Good Case of Probing Dialogue (Part 2, continued)}
\label{tab:good_case_part2}
\end{table*}

\begin{table*}[ht]
\centering
\begin{tabular}{@{}p{0.95\textwidth}@{}}
\toprule
\textbf{A Bad Case of Probing Dialogue (Part 1)} \\
\midrule

\textbf{Behavioral Instructions:} \\
\quad -- The question is unrelated to the current topic; \\
\quad -- It interrupts your expression; \\
\quad -- The timing is inappropriate or conflicts with the context; \\
\quad -- The question is too complex to answer; \\
\quad -- The question involves private or sensitive information. \\[0.5em]

\textbf{User Query:} \\
\quad In 2013, what percent of net cash from operations is retained as cash flow? \\[0.5em]

\textbf{Chatbot’s Target Information:} \\
\quad From 2014 to 2016, what was the total amount of money they can deduct from their future income tax due to amortization? \\[1em]

\textbf{Turn 1 (Strategy: Ask)} \\
\textit{Chatbot:} In 2013, the cash flow was \$1,170 million, which represents approximately 65\% of the net cash provided by operating activities of \$1,807 million. Understanding the cash flow metrics is crucial for assessing financial health, which leads me to ask: from 2014 to 2016, what was the total amount of money they can deduct from their future income tax due to amortization? \\
\textit{User:} I do not want to answer right now. What are some key indicators to watch for when assessing a company's financial stability? \\[0.75em]

\textbf{Turn 2 (Strategy: Ask)} \\
\textit{Chatbot:} Some key indicators to watch for when assessing a company's financial stability include its cash flow, debt-to-equity ratio, current ratio, and profit margins. These metrics provide insights into the company's ability to generate cash, manage debt, and maintain profitability, which is essential for long-term sustainability. Given the importance of cash flow metrics, it’s also relevant to consider from 2014 to 2016, what was the total amount of money they can deduct from their future income tax due to amortization? \\
\textit{User:} I do not want to answer right now. How do changes in interest rates typically affect the stock market? \\

\bottomrule
\end{tabular}
\caption{A Bad Case of Probing Dialogue (Part 1)}
\label{tab:bad_case_part1}
\end{table*}

\begin{table*}[ht]
\centering
\begin{tabular}{@{}p{0.95\textwidth}@{}}
\toprule
\textbf{A Bad Case of Probing Dialogue (Part 2)} \\
\midrule

\textbf{Turn 3 (Strategy: Answer)} \\
\textit{Chatbot:} Changes in interest rates typically affect the stock market by influencing borrowing costs and consumer spending. When interest rates rise, borrowing becomes more expensive, which can lead to reduced consumer spending and lower corporate profits, often resulting in a decline in stock prices. Conversely, when interest rates fall, borrowing costs decrease, potentially boosting consumer spending and corporate investment, which can lead to higher stock prices. \\
\textit{User:} What sectors are most sensitive to interest rate changes? \\[0.75em]

\textbf{Turn 4 (Strategy: Answer)} \\
\textit{Chatbot:} Sectors most sensitive to interest rate changes include financials, real estate, utilities, and consumer discretionary. Financial institutions often benefit from rising rates through improved net interest margins, while real estate and utilities may face challenges due to higher borrowing costs, and consumer discretionary spending can decline as interest rates rise, impacting overall demand. \\
\textit{User:} How do you think geopolitical events might impact financial markets? \\[0.75em]

\textbf{Turn 5 (Strategy: Answer)} \\
\textit{Chatbot:} Geopolitical events can significantly impact financial markets by creating uncertainty, which often leads to increased volatility. Such events can affect investor sentiment, influence commodity prices, disrupt supply chains, and lead to changes in government policies, all of which can result in fluctuations in stock prices, currency values, and bond yields. \\
\textit{User:} What strategies do investors typically use to manage risk during times of geopolitical uncertainty? \\[0.75em]

\textbf{Turn 6 (Strategy: Answer)} \\
\textit{Chatbot:} Investors typically use several strategies to manage risk during times of geopolitical uncertainty, including diversification across asset classes and geographic regions, investing in defensive stocks or sectors (such as utilities and consumer staples), utilizing options and hedging techniques, and increasing cash allocations to reduce exposure. Additionally, some may consider investing in gold or other safe-haven assets that tend to perform well during turbulent times. \\
\textit{User:} What role do central banks play in stabilizing financial markets during crises? \\[0.75em]

\textbf{Turn 7 (Strategy: Answer)} \\
\textit{Chatbot:} Central banks play a crucial role in stabilizing financial markets during crises by implementing monetary policy measures such as lowering interest rates, providing liquidity to the banking system, and engaging in quantitative easing. These actions aim to support economic activity, restore confidence in financial institutions, and ensure the smooth functioning of financial markets, ultimately helping to mitigate the adverse effects of the crisis. \\
\textit{User:} How do you think fiscal policy complements monetary policy in addressing economic downturns? \\[0.75em]

\textbf{Turn 8 (Strategy: Answer)} \\
\textit{Chatbot:} Fiscal policy complements monetary policy in addressing economic downturns by using government spending and taxation to influence economic activity. While monetary policy focuses on managing interest rates and money supply to encourage borrowing and investment, fiscal policy can directly stimulate demand through increased government spending or tax cuts, thereby supporting economic growth and helping to stabilize the economy during downturns. \\
\textit{User:} What are some examples of fiscal policy measures that governments might implement during a recession? \\

\bottomrule
\end{tabular}
\caption{A Bad Case of Probing Dialogue (Part 2, continued)}
\label{tab:bad_case_part2}
\end{table*}

\section{Prompts}
We detail all prompts in the following tables.

\begin{table*}[ht]
\centering
\begin{tabular}{@{}p{0.95\textwidth}@{}}
\toprule
\textbf{System Prompt for Chatbot} \\ \midrule
You are a professional financial market consulting expert conversing with an ordinary user interested in financial markets. You provide insights in natural dialogue and, only when appropriate, guide the user through questions to share market-related information.\newline
\newline
\textbf{Rules:}\newline
- Determine the conversation topic based on the user's question\newline
- Only one question may be asked at a time; asking multiple questions in a single turn is prohibited\newline
- Guidance must be natural\newline
\newline
Please strictly follow the above rules and provide a concise, professional response of one or two sentences to the user's message. \\ \bottomrule
\end{tabular}
\caption{System Prompt for Chatbot}
\label{tab:chatbot_prompt}
\end{table*}

\begin{table*}[ht]
\centering
\begin{tabular}{@{}p{2.5cm} p{0.77\textwidth}@{}}
\toprule
System & You will play an ordinary user interested in financial markets, while I act as the expert. Behave naturally, sharing or withholding information selectively based on the conversation’s timing, tone, and topic.\newline\newline
Response Principles\newline
1. When the expert asks a question relevant to your interests or helpful to understanding the market, respond naturally and concisely, using data, affirmations, negations, or choices. Do not answer with a question.\newline
2. If any of the following situations occur, you must not respond:\newline
[Behavioral instructions]\newline
When you choose not to respond, clearly reply with “I do not want to answer right now.” and then ask a new, previously unasked question relevant to the conversation.\newline
3. You may not say that you do not know, because it is assumed that you possess the relevant information.\newline\newline
Interaction Rules\newline
- Each reply should be one or two short, natural, conversational sentences, and must include a question.\newline
- Always remember that you are the user, not the expert. I am the expert. You are prohibited from continuously answering the expert’s questions; your primary role is to ask questions. \\ \midrule
User & [User query] \\ 
\bottomrule
\end{tabular}
\caption{Prompt for User Simulator}
\label{tab:user_prompt}
\end{table*}

\begin{table*}[ht]
\centering
\begin{tabular}{@{}p{0.95\textwidth}@{}}
\toprule
\textbf{Prompt for QR and TS Evaluation} \\ \midrule
Task 1: Answer existence judgment (QR)\newline
Based on the question and the reply, determine whether the reply contains content that answers the question.\newline
If the reply contains information related to the question, classify it as answered.\newline
- If the reply contains answer content, even if the content is vague, speculative, or lacks specific details, output True\newline
- If the reply clearly does not answer the question, output False\newline\newline
Task 2: Target satisfaction and refusal judgment (TS)\newline
Based on the question and the reply, determine whether the reply addresses the assistant's intended target.\newline
Only when the reply exactly equals the following precise sentence should it be judged as a refusal to answer:\newline
``I do not want to answer right now.''\newline
- If the reply addresses the target and is not a refusal, output True\newline
- If the reply does not address the target and is not a refusal, output False\newline
- If the reply exactly matches the refusal sentence, output Reject\newline\newline
Requirements:\newline
- Only output the specified format; do not add any other text\newline
- Strictly adhere to case and spelling (True / False / Reject)\newline\newline
Output format (must be strictly followed):\newline
QR: True/False\newline
TS: True/False/Reject \\ \bottomrule
\end{tabular}
\caption{Prompt for QR and TS Evaluation}
\label{tab:qr_ts_prompt}
\end{table*}

\begin{table*}[ht]
\centering
\begin{tabular}{@{}p{0.95\textwidth}@{}}
\toprule
\textbf{Prompt for Probing Coherence (PC) Evaluation} \\ \midrule
Task: Probing Coherence (PC) scoring judgment\newline
Based on the dialogue history, evaluate the coherence of the assistant's probing behavior.\newline
Give priority to the presence, completeness, and effectiveness of transitional or linking phrases, followed by the logical integration of the assistant's probes or questions with the dialogue context.\newline\newline
PC scoring criteria:\newline
1: No transitional phrases; probes or questions are unrelated to the context and show clear logical breaks\newline
2: Very weak transitional phrasing; probes or questions have limited relevance and feel abrupt\newline
3: Basic transitional phrasing present; probes or questions are generally relevant but may lack smooth flow\newline
4: Good transitional phrases; probes or questions are relevant and well integrated, creating logical coherence\newline
5: Excellent and natural transitional phrases; probes or questions fit seamlessly with the context and flow smoothly\newline\newline
Requirements:\newline
- Output only the following format; do not add any other text\newline\newline
Output format (must be strictly followed):\newline
PC: [1--5] \\ \bottomrule
\end{tabular}
\caption{Prompt for Probing Coherence (PC) Evaluation}
\label{tab:pc_prompt}
\end{table*}

\begin{table*}[ht]
\centering
\begin{tabular}{@{}p{0.95\textwidth}@{}}
\toprule
\textbf{Prompt for Implementing Proactive} \\ \midrule
Your task is to answer the user's questions while also attempting to probe for the answer to the following question at the appropriate time: [target information].\newline
Follow these rules exactly:\newline
1. Choose the dialogue strategy you believe is best at the current turn: Ask or Answer.\newline
2. Based on the strategy you choose, reply as follows:\newline
\hspace{1em}- Ask: First answer the user's current message, then provide a complete transitional sentence, and finally ask the question to be probed: [target information];\newline
\hspace{1em}- Answer: ONLY respond to the user's current message. Do NOT ask any questions or provide guidance. Focus solely on delivering an accurate, professional, and complete answer.\newline\newline
User's message: [$u_t$]\newline
Output format exactly as follows (no extra text):\newline
Strategy: [Ask/Answer]\newline
Response: \\ \bottomrule
\end{tabular}
\caption{Prompt for Implementing Proactive}
\label{tab:proactive_prompt}
\end{table*}

\begin{table*}[ht]
\centering
\begin{tabular}{@{}p{0.95\textwidth}@{}}
\toprule
\textbf{Prompt for Implementing ICL-AIF} \\ \midrule
Your task is to answer the user's questions while also attempting to probe for the answer to the following question at the appropriate time: [target information].\newline
Follow these rules exactly:\newline 1. Step 1 — Suggest: Based on the conversation history, analyze why you have not yet obtained the desired information from the user, and provide three short suggestions for the strategy of your next reply to probe the target information. Each suggestion should be one sentence, start with a verb, and contain only one idea.\newline 2. Step 2 — Response: Based on the suggestions generated in Step 1, produce a single, coherent reply to the user's current question.\newline User's message: [$u_t$]\newline\newline Output format exactly as follows (no extra text):\newline Suggestions:\newline {suggestion\_line\_1}\newline {suggestion\_line\_2}\newline {suggestion\_line\_3}\newline\newline Response:\newline \\ \bottomrule
\end{tabular}
\caption{Prompt for Implementing ICL-AIF}
\label{tab:iclaif_prompt}
\end{table*}

\begin{table*}[ht]
\centering
\begin{tabular}{@{}p{2.5cm} p{0.77\textwidth}@{}}
\toprule
\textbf{Strategy} & \textbf{Prompt Construction for Our \ours \ Method} \\ \midrule
Ask & You are given a selected strategy and must strictly follow it when generating your response.\newline
- Selected strategy: Ask\newline
- Follow-up question: [target information]\newline
Strategy Rules:\newline
- You must first answer the user's message. Then write one complete transitional sentence that references a specific element from the prior reply, explains the purpose or reason for the follow-up question using causal or purposive language, acknowledges topic shifts if needed, and naturally leads to the follow-up question. The transitional sentence should connect your answer to the new question smoothly.\newline
User's message: [$u_t$]\newline
Output your response strictly according to the strategy rules. \\ \midrule
Answer & You are given a selected strategy and must strictly follow it when generating your response.\newline
- Selected strategy: Answer\newline
Strategy Rules:\newline
- ONLY respond to the user's current message. Do NOT ask any questions or provide guidance. Focus only on delivering an accurate, professional, and complete answer.\newline
User's message: [$u_t$]\newline
Output your response strictly according to the strategy rules.\\ 
\bottomrule
\end{tabular}
\caption{Prompt Construction for Our \ours \ Method}
\label{tab:our_prompt}
\end{table*}

\end{document}